\documentclass[10pt,twocolumn,letterpaper]{article}

\usepackage[pagenumbers]{cvpr} % To force page numbers, e.g. for an arXiv version

\definecolor{cvprblue}{rgb}{0.21,0.49,0.74}
\usepackage[pagebackref,breaklinks,colorlinks,allcolors=cvprblue]{hyperref}
\usepackage{svg}
\usepackage[accsupp]{axessibility}  % Improves PDF readability for those with disabilities.

\usepackage{multirow}

\usepackage{booktabs}
\usepackage[normalem]{ulem}
\usepackage{cuted}
\usepackage{float}
\usepackage{subcaption}

\def\paperID{*****} % *** Enter the Paper ID here
\def\confName{CVPR}
\def\confYear{2026}

\title{{BUSSARD}: Normalizing Flows for Bijective Universal Scene-Specific Anomalous Relationship Detection}

\author{Melissa Schween \quad Mathis Kruse \quad Bodo Rosenhahn\\
Institute for Information Processing, L3S - Leibniz University Hannover\\
{\tt\small schween@tnt.uni-hannover.de}\\
}

\def\method{\emph{BUSSARD}}

\begin{document}
\maketitle
\begin{abstract}
We propose Bijective Universal Scene-Specific Anomalous Relationship Detection (\method), a normalizing flow-based model for detecting anomalous relations in scene graphs, generated from images.
Our work follows a multimodal approach, embedding object and relationship tokens from scene graphs with a language model to leverage semantic knowledge from the real world.
A normalizing flow model is used to learn bijective transformations that map object-relation-object triplets from scene graphs to a simple base distribution (typically Gaussian), allowing anomaly detection through likelihood estimation.
We evaluate our approach on the SARD dataset containing office and dining room scenes.
Our method achieves around 10\% better AUROC results compared to the current state-of-the-art model, while simultaneously being five times faster.
Through ablation studies, we demonstrate superior robustness and universality, particularly regarding the use of synonyms, with our model maintaining stable performance while the baseline shows 17.5\% deviation.
This work demonstrates the strong potential of learning-based methods for relationship anomaly detection in scene graphs.
Our code is available at \url{https://github.com/mschween/BUSSARD}.
\end{abstract}    
\section{Introduction}
\label{sec:intro}
Detecting anomalies in images is not always straightforward, as the definition of what constitutes an anomaly depends on the setting and context of the scene.
For industrial applications, anomalies typically manifest as deformed or missing objects~\cite{dataset:mvtec, dataset:loco-ad}.
The automatic detection of anomalies in surveillance footage and similar domains is gaining research attention, addressing complex data with substantial variability~\cite{dataset:ucf-crime, dataset:shanghaiTec, lai2025scene}.
In this area, the focus shifts from identifying individual defects to understanding entire scenes and evaluating context, with the challenge of filtering what is relevant and necessary to consider.
Most existing approaches focus on individual scene components, particularly human poses, with limited integration of broader contextual information~\cite{humanAD:pose, humanAD:pose-MOT, humanAD:pose-NF}.
However, this human-centric focus neglects important anomaly types, such as objects in inappropriate locations (\eg car on playground) or unusual human-object relationships (\eg child on car).
Recognizing this gap, Lai \etal~\cite{lai2025scene} introduced the novel task of \textit{Scene-Specific Anomalous Relationship Detection (SARD)} along with a dedicated dataset.

\begin{figure}[!t]
  \centering
  \includegraphics[width=1.0\linewidth]{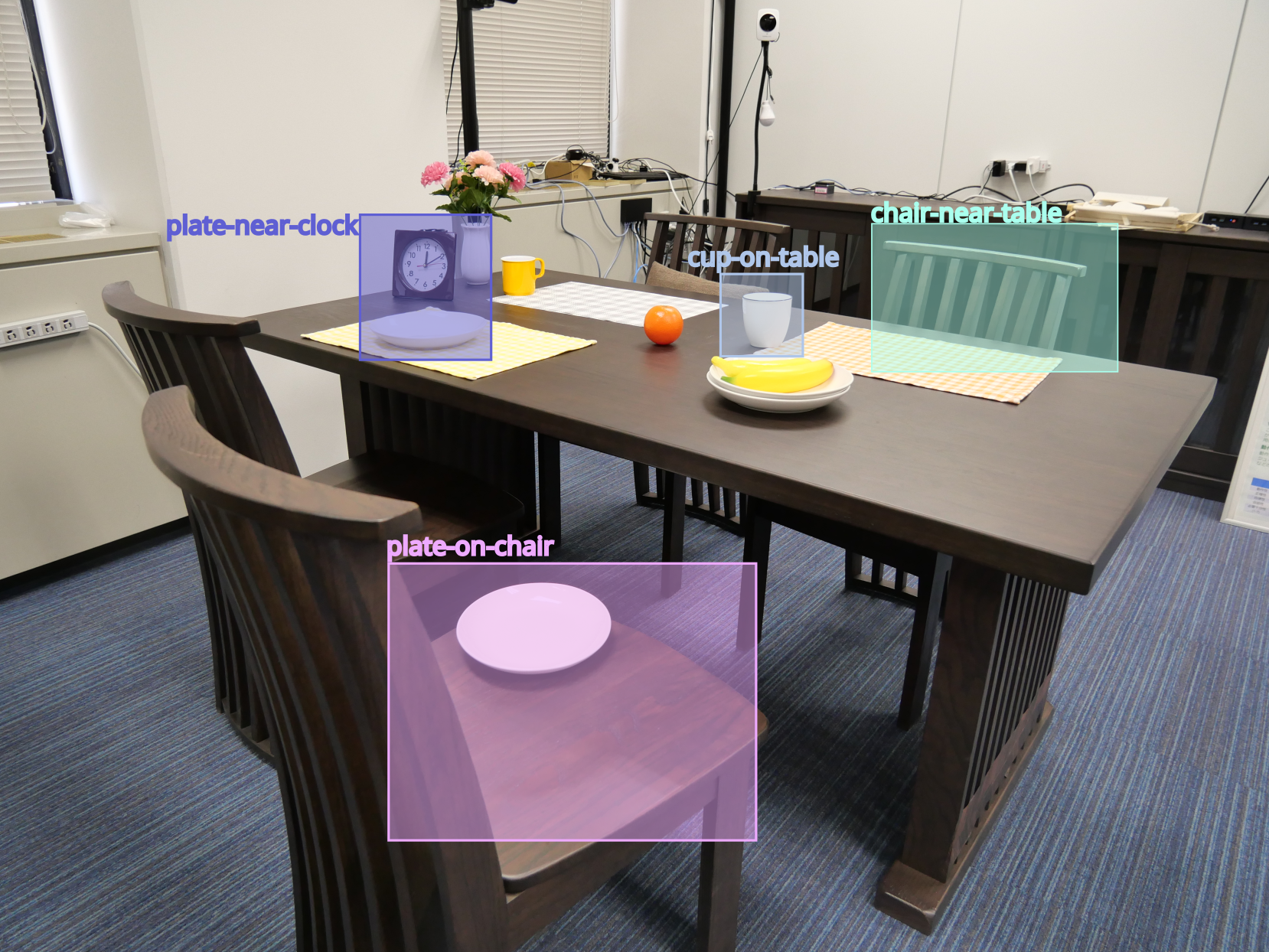}
   \caption{Example image from SARD dataset~\cite{lai2025scene}. A non complete scene graph consists of: `plate-on-chair', `plate-near-clock', `cup-on-table', `chair-near-table'. The anomaly to detect is `plate-on-chair'.}
   \label{fig:sg-example}
\end{figure}

Recent progress in scene graph generation (SGG)~\cite{egtr, univSG, synthVG, sgg-llm2}, accelerates the research of  applying SGGs for anomaly detection (AD) in complex scenes through structured extraction of visual information.
While this combination remains underexplored~\cite{uncondscene2021garg, scene-graph-video-AD, scene-graph-video-AD2}, it offers significant potential by interpreting structured scene representations for more effective anomaly detection.
\cref{fig:sg-example} highlights some object relations that can be extracted from the image using SGG.
More examples with complete scene graphs can be found in the appendix (see \cref{sec:img-sg-example}).

Lai \etal~\cite{lai2025scene} propose a counting-based approach to detect anomalous relationships for the SARD task.
A key limitation of this method is its vulnerability to the long-tail problem, where data exhibits severe imbalance with a few instances appearing frequently while the majority is rare.
This counting-based approach cannot fully compensate for such distributional bias, leading to degraded detection performance.
We present, to the best of our knowledge, the first learning-based solution for SARD.
Our approach makes three core contributions to efficiently approach the task of SARD:

First, we follow a multimodal approach by converting scene graphs to semantic vector representations using a pretrained language model to capture real-world knowledge about object and relationship semantics.
This embedding approach addresses the long-tail distribution of objects and relationships in scene graphs, where many entities appear rarely in training data.
By placing semantically similar words in close proximity within the embedding space, the model can generalize to previously unseen objects.

Second, we leverage normalizing flows for anomaly detection.
Normalizing flows learn bijective transformations that map data to a standard normal distribution, making them well-suited for anomaly detection, as data points that deviate significantly from the normal distribution after transformation can be assigned high anomaly scores based on their log-likelihood.

Finally, we address the dimensionality challenge inherent to normalizing flows.
Since normalizing flows require bijective mappings, they cannot perform dimensionality reduction.
However, language embeddings are typically high-dimensional, which can lead to training instability and computational inefficiency.
To resolve this, we employ an autoencoder that compresses the concatenated triplet embeddings into a lower-dimensional latent space before feeding them to the normalizing flow.

We call our approach \textbf{B}ijective \textbf{U}niversal \textbf{S}cene-\textbf{S}pecific \textbf{A}nomalous \textbf{R}elationship \textbf{D}etection (\method).
The `bijective' refers to the invertible transformations in normalizing flows, while `universal' reflects the model's robustness to vocabulary variations, a key advantage demonstrated in our experiments.
\method{} achieves state-of-the-art performance with an AUROC 10\% higher than the counting-based SARD baseline.

Our work comprises the following contributions:
\begin{itemize}
    \item We present the first learning-based approach for SARD, employing normalizing flows to learn distributions over scene graph relationships.
    \item We demonstrate superior robustness and universality through semantic embeddings, effectively mitigating the long-tail problem that affects counting-based approaches.
    \item We achieve state-of-the-art performance on the SARD dataset, outperforming the baseline by 10\% in AUROC.
\end{itemize}

\section{Related Work}
\label{sec:relWork}
We review related works from the fields of scene graph generation and anomaly detection.

\subsection{Scene Graph Generation (SGG)}
In scene graph generation (SGG) structured information is extracted from images, which is stored in a graph consisting of multiple triplets. 
These triplets are composed of two objects, and a relationship describing the connection between the objects.
A simple example can be seen in \cref{fig:sg-example}, with triplets such as `plate-on-chair' or `plate-near-clock'.

The major approaches for generating scene graphs may be separated into \emph{one-stage} and \emph{two-stage methods}. 
In two-stage generation, objects are detected first, and then the relations between them are predicted in a second independent step~\cite{two-stage1,two-stage2, two-stage3}.
The one-stage approach differs in that both objects and relations are predicted simultaneously.
Models such as RelTR, OED, PGSG, and EGTR fall under this group~\cite{reltr, egtr, one-stage-OED, one-stage-open-vocab}.
EGTR operates by extracting relation graphs from the multi-head self-attention layers of the DETR~\cite{detr} decoder, treating attention queries and keys as subject and object entities~\cite{egtr}.
Overall, numerous diverse approaches are currently emerging to improve SGG, showing the interest in this field~\cite{univSG, synthVG,sgg-llm,sgg-llm2}.
Besides the image-based SGG there is also a subfield that concentrates on generating scene graphs for videos, called Dynamic Scene Graph Generation (DSGG)~\cite{dsgg1,dsgg2,dsgg3,dsgg4}.
Here scene graphs are connected over time to track how they change.
This enables the recognition of dynamic relations like `person-throwing-ball', instead of `person-holding-ball'.

\textbf{Word Embeddings.}
Scene graphs consist of textual triplets, making pretrained word embeddings a natural choice for encoding semantic information.
Word embedding models such as word2vec~\cite{word2vec} and GloVe~\cite{glove_v0, glove_v1} map words to continuous vector spaces where semantic similarity is reflected by proximity.
This property is particularly valuable for addressing the long-tail distribution in scene graphs: semantically similar but lexically different terms (\eg`person' vs. `human', `sitting on' vs. `seated on') are embedded nearby, enabling the model to generalize across rare or unseen vocabulary.

\subsection{Anomaly Detection (AD)}
The goal of anomaly detection (AD) is to identify unusual, anomalous samples of data that fall outside of what may be considered normal data.
However, a prominent challenge is the severe imbalance in available training data, which contains far more normal instances than anomalous ones.
For this reason, most algorithms are semi-supervised, where the models are trained exclusively on normal data and only encounter anomalous data during testing and inference.
Algorithms that are widely used for AD include memory banks, student-teacher networks and normalizing flows, while Foundation Models are also gaining more popularity.

\textit{Memory banks} are used to store a learned representation of normal data and use it as a reference for comparison during inference~\cite{memory-bank1, memory-bank2, IAD-recall}.

\textit{Student-teacher networks} typically use a pretrained network (such as a ResNet~\cite{resnet} trained on ImageNet~\cite{imagenet}) as the teacher, which is typically trained on some auxiliary task.
The student network is optimized to imitate the teacher using only normal data.
During inference, the student may only be able to mimic the teacher on normal data, leading to high errors for anomalous (unseen) patterns.
Thereby, the difference between student and teacher serves as a proxy for an anomaly score.
A larger difference indicates higher anomaly likelihood, since the student produces more varied results for unseen data.
There are various approaches implementing student-teacher networks for AD~\cite{ast, student-teacher1, student-teacher-csad}.

\textit{Foundation Models} are trained on massive datasets and can be applied to a wide range of downstream tasks.
One such task is AD, as their learned understanding of the world helps to identify what might be unusual.
A key advantage is their applicability in \textit{few-shot} or \textit{zero-shot} settings, requiring minimal or no task-specific training.
Therefore, several approaches have been proposed to address AD tasks, but they typically require careful pipeline design tailored to the specific problem or they rely on fine-tuned prompts to overcome challenges~\cite{two-sam-ad, foundationAD:AnomalyCLIP, foundationAD:TrainingFree, foundationAD:WinCLIP}.
Specifically there are multiple approaches using GPT~\cite{openai_chatgpt} for zero-shot AD~\cite{foundationAD:AnomalyGPT, foundationAD:LogiCode, foundationAD:LogicAD}.

\textit{Normalizing Flows} learn a bijective mapping between a simple base distribution and a complex target distribution.
The bijective nature enables their use for applications such as data generation~\cite{nf-application, nf-architecture, nf-architecture2}, by sampling from the base distribution and applying the reverse mapping to the complex distribution.
Additionally, normalizing flows provide exact and tractable log-likelihood evaluation, which is essential for their application to AD, as demonstrated by several existing methods~\cite{IAD-MultiFlow, humanAD:pose-NF, NF-differNet}.

Anomaly detection finds application across diverse domains, with unique challenges and specialized approaches.

\textit{Industrial applications} represent one of the most mature areas, supported by well-established benchmarks such as MVTec AD and VisA~\cite{dataset:mvtec, dataset:mvtec2, dataset:loco-ad, dataset:visa}. These datasets have motivated numerous methods advancing detection capabilities~\cite{ast, IAD-recall, IAD-MultiFlow, dinomaly}. The medical field has similarly developed domain-specific datasets and approaches~\cite{dataset:bmad, medicalAD, medicalAD2}.

\textit{Video-based AD} introduces temporal complexity, requiring methods to track elements over time. This is particularly relevant for surveillance applications, where datasets like ShanghaiTech and UCF-Crime~\cite{dataset:ucf-crime, dataset:shanghaiTec} have driven the development of pose-based approaches, sometimes augmented with contextual information~\cite{humanAD:skeleton, humanAD:pose, humanAD:pose-MOT, humanAD:pose-NF}.

\textit{Graph-based AD} addresses structured relational data, in domains such as attributed networks, social networks and industrial control systems~\cite{graphAD:industrial, graphAD:attributedNet, graphAD:social, graphAD:physics}.
Recent works have begun incorporating spatial context through context graphs~\cite{scene-aware-AD} and spatio-temporal graphs~\cite{graphAD:anoGraph}.

\textbf{Anomaly Detection using Scene Graphs.}
Recent advances in SGG have sparked increased interest in applying scene graphs to anomaly detection, a relatively underexplored area.
Garg \etal~\cite{uncondscene2021garg} focus on generating new scene graphs, primarily for image generation, and demonstrate anomaly detection on their own generated outputs.
However, their method requires node classification for likelihood estimation, necessitating predefined entity classes for all objects, and evaluates generation quality rather than detecting anomalies in observed scene graphs from real images, making it unsuitable as a baseline for our task.
The work of Chen \etal showed the potential of using scene graphs for video-based anomaly detection~\cite{scene-graph-video-AD}.
ComplexVAD a recently published method, also uses scene graphs for anomaly detection in videos and highlights the importance of focusing on interaction anomalies~\cite{scene-graph-video-AD2}. 
The authors of SARD highlight the importance of focusing on anomalous relationships and use a counting based approach for detecting those using scene graphs in images~\cite{lai2025scene}.
SARD is explained in more detail in the following section.

\section{Foundations}
\label{sec:foundations}
This section provides a brief overview of the \textit{Scene-Specific Anomalous Relationship Detection (SARD)} task~\cite{lai2025scene}, its baseline model, and the evaluation metrics employed.
SARD shifts the focus from detecting anomalous objects to identifying context-dependent anomalies by analyzing relationships between objects.
This relationship-based approach captures anomalies that arise from unusual interactions within a scene.

\begin{figure*}[t]
  \centering
  \includegraphics[width=1.0\textwidth]{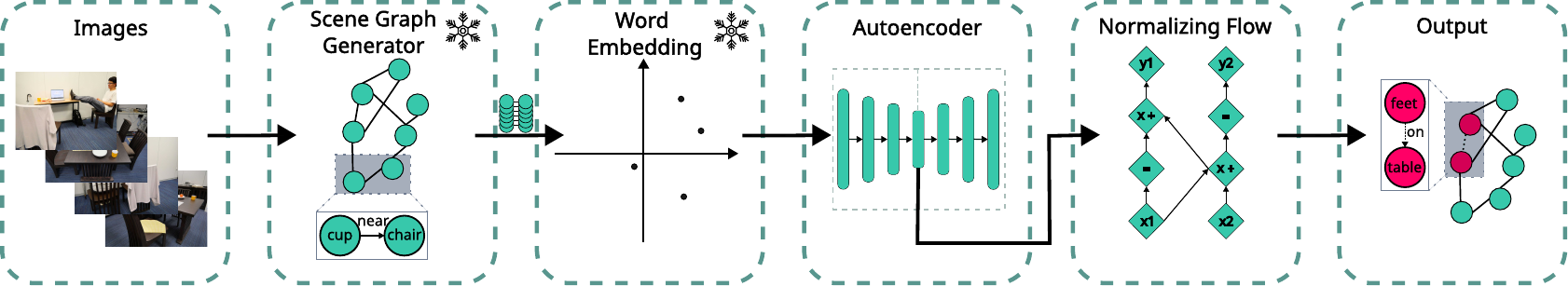}
   \caption{The components of \method. The images are parsed using a pretrained scene graph generator. Each triplet is then encoded using a pretrained word embedding model. The embeddings of the triplets are each concatenated and the dimension is reduced using an autoencoder. In the end, a normalizing flow is used to predict the likelihood of the triplets being anomalous.}
   \label{fig:model-pipe}
\end{figure*}

\subsection{Baseline Method}
\label{subsec:SARD}
We use the AD-model of Lai \etal as baseline~\cite{lai2025scene}.
The authors use EGTR~\cite{egtr} for creating scene graphs for each image to extract the relevant information.
Subsequently the top 30 triplets with the highest confidence for every image are used for further analysis.
Lai \etal do not train a neural network to detect anomalous triplets but instead employ an enhanced counting-based approach, where the triplets that appear least often are considered to be an anomaly.
For this, scene graphs from multiple images are grouped in subgroups with a specified size.
Each subgroup, that can have 11, 21 or 31 elements, consists of one anomalous image with the remainder filled with normal images.
The normal images appear in multiple subgroups.

Scene graphs of each subgroup are grouped separately and the triplets that appear the least, are considered to be anomalies.
For enhancing the counting-based approach, they include the concepts \textit{rarity adjustment}, \textit{noisy normal relationship reduction} and \textit{soft counting}.
Rarity adjustment is used to balance the confidence score bias of the SGG model, as those models tend to assign anomalous relationships lower confidence scores which can lead to an overshadowing through the normal relationships.
Noisy normal relationship reduction is added to reduce the likelihood of classifying infrequent, but normal relationships as anomalies.
The last enhancement employs soft counting to improve the robustness of object relationship representations, as it enables the consideration of multiple likely relationships for each object pair.
These enhancements increase the performance of their model, however improving the biased confidence score, or reducing the influence of rare normal relationships does not solve the problem of unbalanced data that is typically found in AD.
Similarly, efforts to increase robustness through considering multiple relationships do not fully enable accurate performance in an open-world setting.

\subsection{Metrics}
A common metric to evaluate the performance of anomaly detection frameworks is Area Under the Receiver Operating Characteristic Curve (\textit{AUROC}).
AUROC is commonly used as it is scale- and threshold-invariant, which helps to measure the quality of prediction rankings, irrespective of the classification threshold chosen.
Conceptually, AUROC corresponds to the area under the curve of the true positive rate (TPR) as a function of the false positive rate (FPR) with
\begin{equation} \label{eq:auroc}
\text{AUROC} = \int_0^1 \text{TPR}(u) \, d(\text{FPR}(u)),
\end{equation}
where both rates are evaluated at varying thresholds $u$.

The authors of SARD introduce an additional metric, \textit{AUC-Recall@k}, to assess ranking quality in the top positions~\cite{lai2025scene}.
This metric combines the intuition of \textit{Recall@k}, measuring how many anomalies appear among the top-$k$ ranked samples, with the concept of integrating over the recall curve.
Following the protocol of SARD, 30 triplets with the highest confidence from the SGG are used.
Let $\mathcal{T}$ denote the set of all used triplets extracted from the dataset (or from a particular image subgroup, similar to SARD).
For each triplet $T\in\mathcal{T}$, let $a\in\mathbb{R}$ denote its anomaly score, and define the ground truth anomaly indicator
\begin{equation}
    y = \begin{cases}
        1 & \text{if }T\text{ is anomalous} \\
        0 & \text{otherwise}.
    \end{cases}
\end{equation}
Let $N$ be the total number of anomalous triplets (assume $N>0$), and let $r(T)$ denote the rank of triplet $T$ when all triplets are sorted in descending order by their anomaly scores.
Then, Recall@$k$ is defined as
\begin{equation} \label{eq:recall_at_k}
\text{Recall@}k = \frac{1}{N}\sum_{T\in\mathcal{T}} \mathbf{1}\{ r(T) \le k\} \cdot y,
\end{equation}
where $\mathbf{1}\{\cdot\}$ is the indicator function.
To aggregate Recall@$k$ across a range of $k$ values (following the SARD protocol), we define AUC-Recall@$k$ as the normalized discrete area under the Recall@$k$ curve for integer $k\in{k_{\min},\dots,k_{\max}}$
\begin{equation} \label{eq:auc_recall_k}
\text{AUC-Recall@}k = \frac{1}{k_{\max}-k_{\min}+1} \sum_{k=k_{\min}}^{k_{\max}} \text{Recall@}k.
\end{equation}
In our experiments, we set $k_{\min}=1$ and $k_{\max}=100$, following the setting of SARD. A higher AUC-Recall@$k$ indicates that anomalous triplets are ranked closer to the top of the list.
Since Lai \etal evaluate ranking quality within contextual subgroups, AUC-Recall@$k$ is also computed separately for each image subgroup.

AUROC thus captures global discrimination ability across all thresholds, while AUC-Recall@$k$ focuses on the ranking performance of the most anomalous triplets within the top-$k$ positions.

\section{Method}
\label{sec:method}
An overview of \method{}'s architecture is presented in \cref{fig:model-pipe}.
The model comprises four sequential steps.
First, a pretrained SGG model extracts scene graphs from images, providing triplets for analysis.
Second, a pretrained word embedding model converts the object and relationship tokens to semantic vector representations, producing rich embeddings.
Afterwards, an autoencoder compresses these features to lower-dimensional latent vectors.
This compression is essential for the normalizing flow, as these models require matching input and output dimensions due to their bijective nature and exhibit training instability with high-dimensional inputs.
The compressed latent vectors are then passed to the normalizing flow, which assigns anomaly scores to triplets.
The SGG model and word embedding models are kept frozen to preserve their learned general semantic knowledge from pretraining.

\subsection{Scene Graph Generator}
The SGG model takes images $I$ as input and generates scene graphs $G = (V,E)$, where $V$ represents nodes and $E$ represents edges.
We consider subsets of the scene graphs in the form of triplets $T_{i,j} = (o_i,p_{i,j},o_j)$ which consist of two nodes (objects) and one directed edge (relationship), where $o_i, o_j \in V$ and $p_{i,j} \in E$.
Thus the triplet $T_{i,j}$ describes the relationship $p_{i,j}$ from object $o_i$ to object $o_j$.

\subsection{Word Embedding}
We employ the pretrained word embedding model GloVe~\cite{glove_v1} to map object and relationship tokens to vector representations. This leverages learned semantic knowledge from large-scale pretraining, encoding meaningful relationships between words in the embedding space.
% All tokens are translated into vectors of uniform embedding size $d$.
Each object is mapped to a vector $ \mathbf{v}_{o_i} \in \mathbb{R}^d $ and each relationship $p_{i,j}$ is mapped to a vector $\mathbf{v}_{p_{i,j}} \in \mathbb{R}^d$.
Afterwards, the vectors are concatenated to form a unified triplet representation with
\begin{equation} \label{eq:word-embed}
    \mathbf{t}_{i,j} = \text{concat}(\mathbf{v}_{p_{i,j}}, \mathbf{v}_{o_i}, \mathbf{v}_{o_j}) \in \mathbb{R}^{3d} .
\end{equation}
In the following, we drop the indices $i$ and $j$ for simplicity, referring simply to triplets $\mathbf{t}$ when the context is clear.

\subsection{Autoencoder}
An autoencoder (AE) compresses concatenated triplet vectors into a lower-dimensional latent space.
It consists of an encoder $f_{\text{enc}}: \mathbb{R}^{3d} \rightarrow \mathbb{R}^{d_z}$ and a decoder $f_{\text{dec}}: \mathbb{R}^{d_z} \rightarrow \mathbb{R}^{3d}$, mapping each triplet $\mathbf{t}$ to a latent vector $\mathbf{z} = f_{\text{enc}}(\mathbf{t})$ and reconstructing it as $\hat{\mathbf{t}} = f_{\text{dec}}(\mathbf{z})$, where $d_z < 3d$.
The AE is trained on normal data by minimizing the mean squared error (MSE) reconstruction loss
\begin{equation}
    \mathcal{L}_{\text{AE}} = \frac{1}{|\mathcal{T}|} \sum_{T \in \mathcal{T}} \|\mathbf{t} - \hat{\mathbf{t}}\|^2.
\end{equation}
After training the AE in a first stage, only the encoder with frozen weights is used to generate latent representations for the normalizing flow to enable an efficient computation with dense information.

\subsection{Normalizing Flow}
The latent vectors are passed to a normalizing flow to compute anomaly scores for all triplets.
Let $\mathbf{u} \in \mathbb{R}^{d_z}$ denote samples from a simple base distribution, which is commonly a standard Gaussian $\mathbf{u} \sim \mathcal{N}(0, I)$ with density $p_\textbf{u}(\textbf{u})$ and let $\mathbf{z} \in \mathbb{R}^{d_z}$ denote samples from the target distribution.
The normalizing flow $f_{\text{flow}}: \mathbb{R}^{d_z} \rightarrow \mathbb{R}^{d_z}$ models an invertible transformation that maps the target distribution to the base distribution with $\mathbf{u} = f_{\text{flow}}(\mathbf{z})$.

The density $p_\mathbf{z}(\mathbf{z})$ can be estimated using the change of variables formula as
\begin{equation}
    p_\mathbf{z}(\mathbf{z}) = p_\mathbf{u}\left(f_\text{flow}\left(\mathbf{z}\right)\right) \left|\det \left( \frac{\partial f_\text{flow}}{\partial \mathbf{z}}\right) \right|.
\end{equation}
Normalizing flows require invertible transformations with efficiently computable Jacobian determinants.
Various architectures have been proposed to satisfy these requirements~\cite{nice-NF, glow-NF, nf-architecture}, with RealNVP~\cite{realNVP} being one of the most widely used approaches.
RealNVP employs coupling layers with affine transformations, which are trivially invertible and have tractable Jacobian determinants  due to their triangular structure, enabling scalable density modeling in high-dimensional spaces.
Each coupling layer splits the input $\mathbf{z}_\text{in} \in \mathbb{R}^{d_Z}$ into two parts $\mathbf{z}_{\text{in},1}\in \mathbb{R}^{d_z/2}$ and $\mathbf{z}_{\text{in},2} \in \mathbb{R}^{d_z/2}$.
The first part remains unchanged while the second part undergoes an affine transformation conditioned on the first part, leading to the equations
\begin{align}
    \mathbf{z}_{\text{out},1} &= \mathbf{z}_{\text{in},1}\\
    \mathbf{z}_{\text{out},2} &= \mathbf{z}_{\text{in},2} \odot \text{exp} (s(\mathbf{z}_{\text{in},1})) + t(\mathbf{z}_{\text{in},1}),
\end{align}
where $s$ and $t$ are deep neural networks computing the scale and translation parameters respectively.
Multiple coupling layers are stacked with alternating permutations between layers to ensure all dimensions are transformed, yielding a flexible and tractable bijective function.

The anomaly score $a$ for each triplet $T$ is computed as the negative log-probability
\begin{equation} \label{eq:nll}
    a = -\log p(\mathbf{z}) = -\log p(\mathbf{u}) - \log \left| \det \frac{\partial f_{\text{flow}}}{\partial \mathbf{z}} \right|,
\end{equation}
where higher scores indicate higher anomaly levels.
For numerical stability, triplets with undefined scores (NaN or Inf) are assigned the maximum observed anomaly score.
The normalizing flow is trained exclusively on normal data with maximizing the negative log-likelihood loss given by
\begin{equation} \label{eq:flow_loss}
    \mathcal{L}_{\text{flow}} = -\frac{1}{2}\|\mathbf{u}\|^2_2 + \log \left| \det \frac{\partial f_{\text{flow}}}{\partial \mathbf{z}} \right|.
\end{equation}

\section{Experiments}
\label{sec:experiments}
Our approach is evaluated on the SARD dataset and compared to what is, to the best of our knowledge, the only available baseline.
We also tested \method{} in a cross-dataset setting, where the model is trained on a subset of the MIT indoor dataset (MIT-67)~\cite{dataset:indoor} (for more details see appendix \cref{sec:mit-67}).
Several ablation studies are conducted to show the robustness of \method{} and underline our design choices.

\subsection{SARD Dataset}
The SARD dataset contains images of two indoor scenes for anomaly detection, specifically created for the SARD task~\cite{lai2025scene}.
The first setting is a dining room scene with 60 normal and 54 anomalous images and the second scene consists of 60 normal and 63 anomalous images.
All images are labeled according to whether they contain an anomaly or represent normal data.
Additionally, the ground truths of anomalous images contain a description of the anomalous triplet.
In each image only one anomaly can be found, but there can be varying valid descriptions for the same anomaly like `cup-on-plate' and `plate-under-cup'.

\begin{figure}[t]
  \centering
  \includegraphics[width=1.0\linewidth]{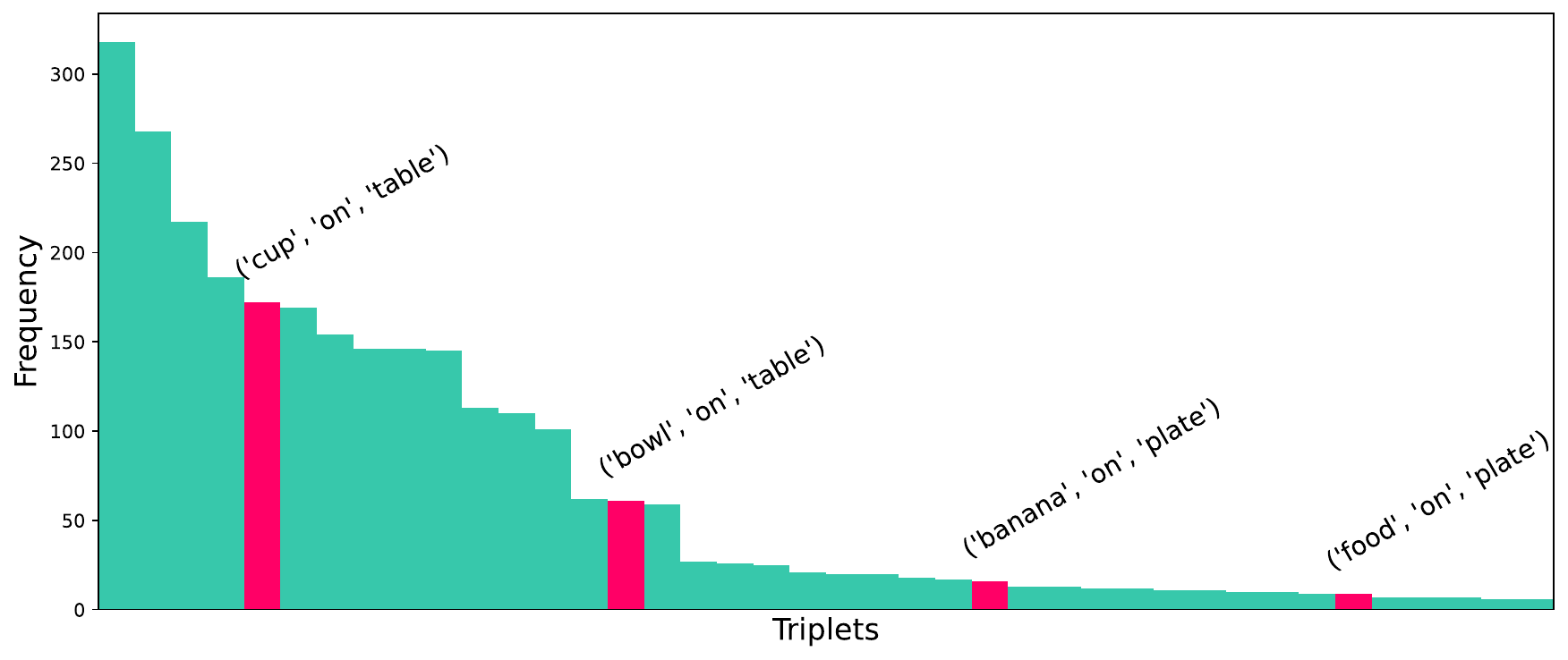}
  \caption{The 40 most frequent triplets of the \textbf{dining room} scene. The labels belong to the highlighted bars, showing example triplets.}
   \label{fig:data_sg_dining_room}
\end{figure}

\textbf{Scene Graph Dataset Analysis and Adaption.} \label{sec:sg-data-analysis}
The images in the dataset are converted to scene graphs using EGTR~\cite{egtr}.
Following the protocol suggested by SARD, the top 30 triplets are used for further analysis, while removing triplets with minor objects as defined by SARD~\cite{lai2025scene} using ConceptNet~\cite{ConceptNet}.
The top 30 triplets of all scene graphs contain approximately 145 unique triplets for the office scene, while the dining room scene contains about 80 unique triplets.
An overview of the distribution of the 40 most common triplets in the dining room scene can be found in~\cref{fig:data_sg_dining_room}, while the distribution of the office scene can be found in the appendix~(\cref{fig:data_sg_office}).
The distribution exhibits the long-tail problem characteristic, as a few triplets are highly frequent while the majority appear very infrequently.
This is why the counting-based approach of Lai \etal has weaknesses, as many normal triplets appear infrequently.

An analysis of all loaded triplets reveals that the anomalous triplet is not always within the top 30 triplets.
This occurs as EGTR assigns lower confidence scores to actual anomalies, causing them to be ranked below the top 30 threshold and are thus excluded from the dataset.
This is problematic since ground truth anomalies that are not present in the extracted triplets cannot be detected by any method.
To ensure fair evaluation, these missing ground truth anomalous triplets are added in a preprocessing step.
For comparability, the baseline SARD method is re-run with this corrected data (SARD-c) in addition to reporting their original results (SARD-o). 

The authors of SARD used and evaluated the performance within subgroups of sizes 11, 21 and 31.
Since \method{} needs to be trained (unlike the SARD model, which operates by counting directly), we use a training/test split of 80/20 for the normal data. This reduces the amount of normal data available for testing.
Consequently, we adopt the subgroup size of 11, pairing each anomalous image with ten normal images from the test set. 

\subsection{Implementation Details}
For converting the images to scene graphs, we use the model EGTR~\cite{egtr}, which is pretrained on Visual Genome~\cite{dataset:visual-genome}. 
The word embedding model GloVe~\cite{glove_v1} is used to translate each word into a vector with dimension $d = 300$.
The three vectors building one triplet are concatenated to one vector $\mathbf{t}$ with dimension $3d = 900$.
The SGG and word embedding models are not trained or finetuned, but instead use frozen pretrained weights.

For dimensionality reduction, an autoencoder is pretrained separately for each scene using only normal training data from the SARD dataset.
The $3d$-dimensional input is passed through four fully connected layers with ReLU activations, with the latent dimension being $d_z = 512$.
Ablation studies (see \cref{sec:design-study}) demonstrate that this latent dimension yields optimal performance.
The decoder mirrors the architecture in reverse.
The autoencoder is trained using the Adam optimizer~\cite{adam} with learning rate $\eta = 0.001$ for 100 epochs, minimizing $\mathcal{L}_{\text{AE}}$.
Only the encoder is used with frozen weights for subsequent anomaly detection, after the training of the AE.

We implement the normalizing flow using three RealNVP coupling layers~\cite{realNVP}.
The first two layers use alternating mask patterns, while the third uses a half mask pattern.
The $s$- and $t$-networks of the coupling layers consist of multi-layer perceptrons (MLPs) that each have three fully connected layers with ReLU activations and a hidden dimension of 128.
The normalizing flow is trained exclusively on normal data for 1000 epochs using the AdamW optimizer~\cite{adamw} and starting with an initial learning rate of $\eta = 10^{-4}$, while maximizing $\mathcal{L}_{\text{flow}}$.
A plateau scheduler is employed to adaptively adjust the learning rate, by reducing the learning rate by a factor of 0.8 when the loss plateaus for 30 consecutive epochs, with a minimum learning rate of $10^{-7}$.
\method{} is trained with ten different random seeds, while SARD is executed once as the counting-based method does not involve stochastic training.

\subsection{Evaluation}
For evaluation, we apply \method{} to the corrected SARD dataset and compare it primarily against SARD-c, ensuring both methods are evaluated on the same corrected scene graphs (see~\cref{sec:sg-data-analysis}).
The original SARD results (SARD-o) are also provided for completeness.
The AUROC and AUC-Recall@k values for each scene are presented in~\cref{tab:main-results}.

\method{} outperforms SARD-o and SARD-c by a large margin in AUROC, achieving 97.85\% and 95.29\% on the dining room and office scenes, respectively, with an average improvement of 10.07\% across both scenes.
\method{} also improves ranking quality, with an average 4.11\% improvement in the AUC-Recall@k metric compared to SARD-c.
Notably, when comparing to the original implementation SARD-o, our method has improved in average 24.21\% in the AUC-Recall@k metric.
A qualitative analysis of a few selected examples is provided in the appendix (see \cref{sec:qualitative-analysis}).

\textbf{Computational Efficiency.}
Beyond detection performance, \method{} demonstrates superior computational efficiency.
Experiments conducted on an NVIDIA RTX 3090 show that, despite SARD-c's simpler counting-based implementation, our approach is on average five times faster.

\begin{table}[t]
\centering
\caption{AUROC and AUC-Recall@k on the SARD dataset. SARD-o refers to the original baseline results with incomplete data (some ground truth anomalies missing from top 30 triplets). SARD-c refers to the baseline re-run with corrected data where all ground truth anomalous triplets are present. BUSSARD is evaluated on the corrected data. Both baselines are included for transparency.
\method{} was executed with ten different seeds and the average and standard deviation are provided. The best performing model is \textbf{highlighted}.}
\label{tab:main-results}
\resizebox{\linewidth}{!}{
\begin{tabular}{l|l||l|l}
    \toprule
    Scene & Model & AUROC ($\uparrow$) & AUC-Recall@k ($\uparrow$)\\
    \midrule
    \multirow{3}{*}{\begin{tabular}[c]{@{}c@{}}Dining\\Room\end{tabular}} & SARD-o & 90.30 & 69.48 \\
     & SARD-c & 90.50 & 91.50\\
     & \method{} \textbf{(ours)} & \textbf{97.85} \scriptsize{$\pm$ 0.6} & \textbf{92.85} \scriptsize{$\pm$ 1.52}\\
    \midrule
    \multirow{3}{*}{Office} & SARD-o & 81.80 & 64.52\\
    & SARD-c & 82.50 & 82.70\\
    & \method{} \textbf{(ours)} & \textbf{95.29} \scriptsize{$\pm$ 1.65} & \textbf{89.57} \scriptsize{$\pm$ 3.46}\\
    \bottomrule
\end{tabular}
}

\end{table}

\subsection{Ablation Studies}
We conduct several ablation studies with special focus on showing robustness, as well as justifying design decisions.

\begin{figure}[t]
  \centering
  \includegraphics[width=0.8\linewidth]{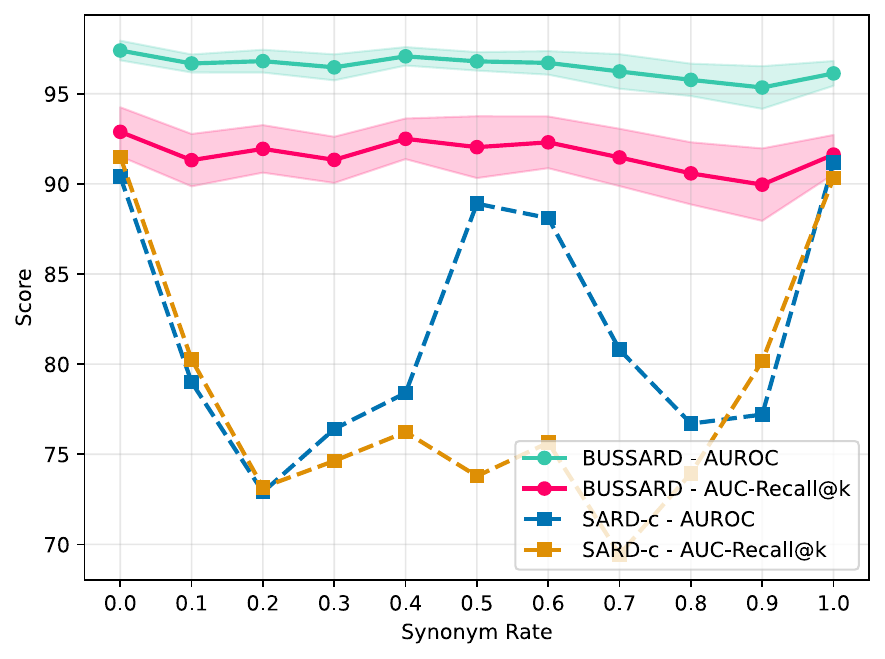}
   \caption{Ablation results with AUROC~($\uparrow$) and AUC-Recall@k~($\uparrow$) of \method{} and SARD-c for different synonym rates for the \textbf{dining room} scene.
   The synonym rate represents the probability of substituting words using synonym mappings.
   For \method{} the dots represent the average results after running with ten different seeds, and the shaded area visualizes the corresponding standard deviation.
   SARD-c was run only once for each rate as the calculation is deterministic.
  }
   \label{fig:eval-mut-dining-room}
\end{figure}

\subsubsection{Robustness}

\textbf{Synonyms.}\label{sec:synonym-exp}
To demonstrate universality in an open-vocabulary setting, we conduct a study where four chosen words are replaced with similar words (see \cref{sec:app-synonym} in appendix for more details).
The replaced words were chosen to be common, while their synonyms are not yet present, making the effect of synonym substitution clearly measurable.
For understanding the influence of the synonyms, words are replaced with different probabilities, here called \textit{synonym rate}.
All ground truths, with triplets containing at least one of the chosen words were modified by adding the triplets with the synonyms as well.

The results of \method{} and SARD-c are shown in \cref{fig:eval-mut-dining-room} for the different synonym rates of the dining room scene. Similar results of the office scene can be found in the appendix \cref{fig:eval-mut-office}.
Most notably, our model remains stable across all synonym rates with around 2\% deviation.
In contrast, the results of SARD-c show significantly stronger deviations with a maximum deviation of around 17.5\%.

We additionally conduct experiments adding \textit{Gaussian noise} to the word embeddings to simulate slight word variances, confirming robustness to small variations.
Further details are provided in the appendix (see \cref{sec:gauss-noise-abl}).

Overall, the synonym experiment and the noise robustness highlight the more universal nature of \method{}, which maintains stable performance across vocabulary variations, in contrast to the less robust enhanced counting approach of SARD-c.

\textbf{Multi-Scene and Cross-Dataset Evaluation.}
To show the robustness of our approach, we provide the results of two additional experiments, summarized in \cref{tab:cross-dataset-ablation-results}.
In both experiments, training is performed on the combined dining room and office scenes, and evaluation is conducted per scene on the SARD test set.
In the first experiment, we use MIT-67 as the training source, and use the combined SARD scene training sets for the second experiment.
\cref{tab:cross-dataset-ablation-results} shows that training on both SARD scenes improves office performance with only a small dining room decline.
Using MIT-67 reveals strong cross-dataset performance on the dining room scene ($92\%$ AUROC), whereas the office scene suffers from a serious decline ($60\%$ AUROC).
We attribute this to the distribution shift between the MIT-67 dataset and SARD's office scenes.

\begin{table}[t]
\centering
\caption{AUROC and AUC-Recall@kk
k comparing multi-scene training on MIT-67 and SARD, evaluated per scene on the SARD test set.}
\resizebox{0.92\linewidth}{!}{
\begin{tabular}{l|l||l|l}
    \toprule
    Train Set & Test Scene & AUROC ($\uparrow$) & AUC-Recall@k ($\uparrow$)\\
    \midrule
    \multirow{2}{*}{SARD} & Dining Room & 97.46 \scriptsize{$\pm$ 0.59} & 91.60 \scriptsize{$\pm$ 2.11} \\
    & Office & 96.73 \scriptsize{$\pm$ 2.06} & 88.53 \scriptsize{$\pm$ 7.31} \\
    \midrule
    \multirow{2}{*}{MIT-67} & Dining Room & 92.45 \scriptsize{$\pm$ 2.50} & 74.79 \scriptsize{$\pm$ 7.93} \\
    & Office & 60.03 \scriptsize{$\pm$ 4.45} & 20.70\scriptsize{$\pm$ 5.63} \\
    \bottomrule
\end{tabular}
}
\label{tab:cross-dataset-ablation-results}
\end{table}

\textbf{Encoding Independence.}
Using RelTR~\cite{reltr} as the scene graph generator yields similar performance (see appendix for more details \cref{sec:encoding-independence}), confirming independence from a specific generator.
Similarly, substituting GloVe~\cite{glove_v1} with stronger sentence embedding models~\cite{word-embed-miniLM, word-embed-mpnet, word-embed-yang2025qwen3} confirms embedding model independence.
More details are provided in the appendix (see \cref{sec:encoding-independence}).

\begin{figure}[b]
  \centering
  \includegraphics[width=0.8\linewidth]{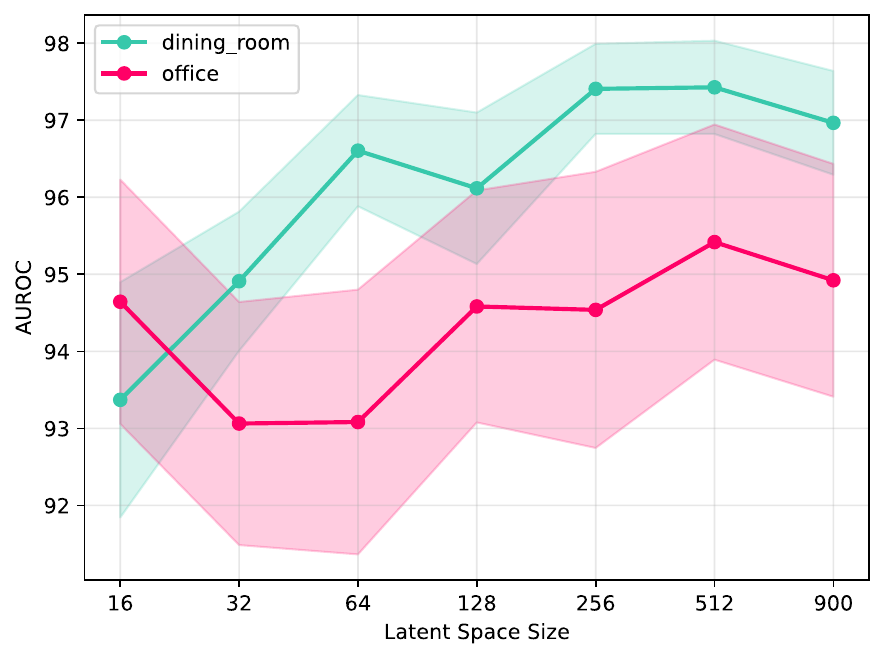}
   \caption{Ablation results with the AUROC ($\uparrow$) for different latent space dimensions of the autoencoder.}
   \label{fig:eval-latent-auroc}
\end{figure}

\begin{table}[t]
\centering
\caption{Ablation study results for the \textbf{dining room} scene. All experiments were executed with ten different seeds and the average results are listed with their standard deviation. $d_z$ is the input dimension to the normalizing flow.}
\label{tab:ablation-dining-room}
\resizebox{\linewidth}{!}{
\begin{tabular}{l|l||c|c}
    \toprule
    Experiment & $d_z$ & AUROC ($\uparrow$) & AUC-Recall@k ($\uparrow$)\\
    \midrule
    Feature Sum & 128 & 91.22 \scriptsize{$\pm$ 2.49} & 78.88 \scriptsize{$\pm$ 4.59}\\
    Feature Mult & 128 & 96.48 \scriptsize{$\pm$ 0.83} & 90.02 \scriptsize{$\pm$ 2.11}\\
    Node Only & 512 & 95.20 \scriptsize{$\pm$ 1.66} & 89.33 \scriptsize{$\pm$ 3.32}\\
    No AE & 900 & 85.44 \scriptsize{$\pm$ 2.33} & 63.40 \scriptsize{$\pm$ 4.78}\\
    \method{} (ours) & 512 & \textbf{97.85} \scriptsize{$\pm$ 0.6} & \textbf{92.85} \scriptsize{$\pm$ 1.52}\\
    \bottomrule
\end{tabular}
}
\end{table}

\subsubsection{Design Study} \label{sec:design-study}
We perform a design study to assess the influence of individual components and justify our design choices.
\\
\par
\textbf{Autoencoder Latent Dimensions}
Since normalizing flows are sensitive to input dimensions~\cite{influence-dimension-nf}, we conduct experiments with varying latent dimension sizes to validate our hyperparameter choice.
The AUROC results across different latent dimensions are presented in \cref{fig:eval-latent-auroc}.
It shows that smaller latent dimensions yield inferior performance compared to larger ones, justifying our choice of $d_z = 512$.
Notably, performance drops again when the latent dimension approaches the input dimensionality (here $d_z=900$).
The AUC-Recall@k results, which exhibit similar behavior, can be found in the appendix (see \cref{fig:eval-latent-auc}).

\textbf{Feature Aggregation Methods.} 
In \method{}, the feature vectors of each word in a triplet are concatenated.
To evaluate whether a more compact representation could be suitable, we performed two experiments that reduce the representation size.
The first variant sums all feature vectors instead of concatenating them, resulting in a three times smaller autoencoder input dimension of $d=300$ (called \textit{Feature Sum}).
Since the chosen latent dimension exceeds this input dimension, the autoencoder is adjusted to compress vectors to $d_z = 128$.
The second variant applies element-wise multiplication of the vectors instead of summation, called \textit{Feature Mult}.
The results of the dining room can be seen in \cref{tab:ablation-dining-room} and the office results are in the appendix (see \cref{tab:ablation-office}).
Both experiments yield inferior results, with Feature multiplication performing closest to \method.

\textbf{Normalizing Flow with only Object Encoding.}
We evaluate the contribution of relationship features by testing a variant that concatenates only object embeddings, called \textit{Node Only} (see \cref{tab:ablation-dining-room} and appendix \cref{tab:ablation-office}). 
The resulting performance drop confirms that incorporating all triplet components, both objects and their relationship, is crucial for effective anomaly detection.

\textbf{Remove Autoencoder.}
To validate the autoencoder's necessity, we tested a variant without it (\textit{No AE}). The results in \cref{tab:ablation-dining-room} and appendix \cref{tab:ablation-office} show that this variant performs significantly worse than all other tested configurations.
This underscores the critical role of the autoencoder in enabling effective normalizing flow performance.

\section{Conclusion}
\label{sec:conclusion}
We propose the first model using a normalizing flow for solving the SARD task.
By using a word embedding model in combination with an autoencoder we are able to create a rich embedding of the triplets that help the normalizing flow to achieve state-of-the-art results on the SARD dataset.
Simultaneously we are able to show the universality of \method{}, improving the applicability to open-word settings.
For future work, the approach could be applied to videos by leveraging DSGG. Additionally large language models could be included to generate textual feedback.

\clearpage
\small{\paragraph{Acknowledgements.} This work was supported by the MWK of Lower Saxony within Hybrint (VWZN4219) and LCIS (VWZN4704), the Deutsche Forschungsgemeinschaft (DFG) under Germany’s Excellence Strategy within the Cluster of Excellence PhoenixD (EXC2122) and Quantum Frontiers (EXC2123), the European Union under grant agreement no. 101136006 – XTREME.
{
    \small
    \bibliographystyle{ieeenat_fullname}
    \bibliography{main}
}

\clearpage
\setcounter{page}{1}
\maketitlesupplementary

\section{Additional Information to the Scene Graph Dataset}\label{sec:appendix_data_SG}
\Cref{fig:data_sg_office} shows the frequency distribution of the 40 most frequent triplets of the office scene.
\begin{figure}[t]
  \centering
  \includegraphics[width=1.0\linewidth]{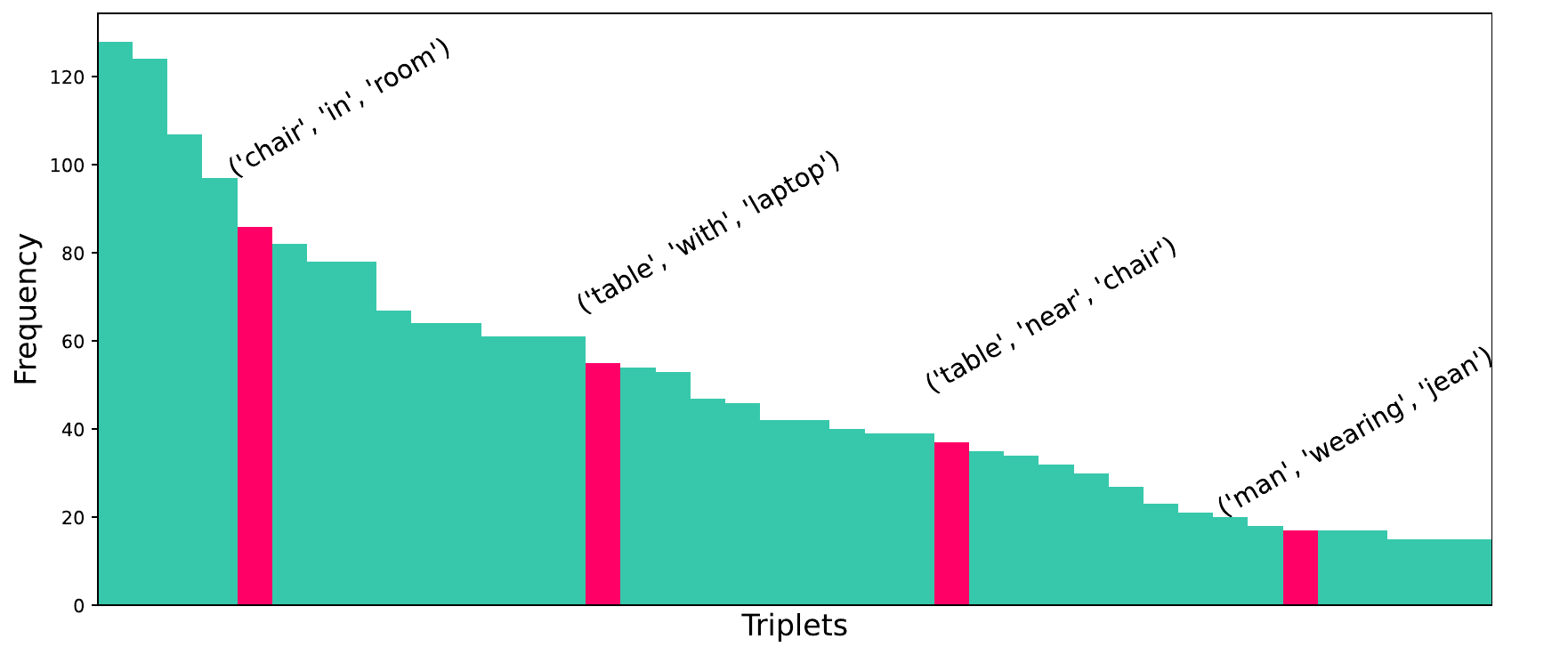}
   \caption{The 40 most frequent triplets of the \textbf{office scene}. The labels belong to the highlighted bars, showing example triplets.}
   \label{fig:data_sg_office}
\end{figure}

\section{MIT-67 Dataset} \label{sec:mit-67}
The MIT-67 dataset consists of various indoor scenes, originally designed for indoor scene recognition~\cite{dataset:indoor}.
Among the overall 67 scenes, there are office and dining room scenes, with each having 109 and 274 samples per scene.
The dataset contains no anomalies and has no specific labels.

\section{Additional Robustness Results}
In this section we provide the additional results for the robustness experiments.

\subsection{Synonym Experiment Details}\label{sec:app-synonym}
The list of the original words and their chosen corresponding synonyms, used for the experiment, are: \textit{table $\rightarrow$ surface}, \textit{chair $\rightarrow$ stool}, \textit{laptop $\rightarrow$ notebook} and \textit{plate $\rightarrow$ dish}.

\Cref{fig:eval-mut-office} shows the results of the synonym experiments for the office scene.
The plateau at 50\% synonym rate occurs since, the original word and its synonym both appear with similar frequency; consequently, when counting, neither word is sufficiently rare to be falsely classified as anomalous.

\begin{figure}[t]
  \centering
  \includegraphics[width=0.8\linewidth]{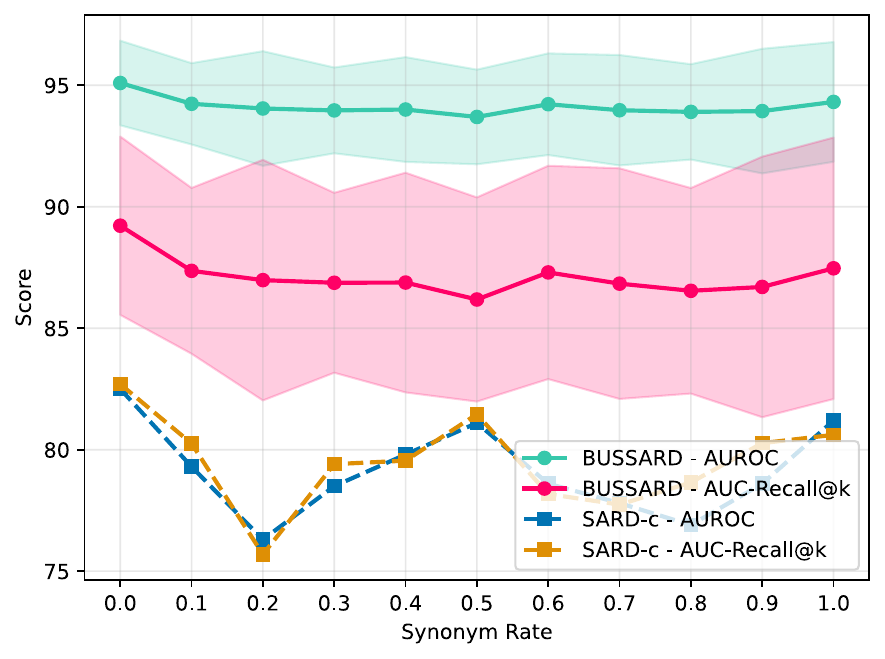}
   \caption{Ablation results with AUROC~($\uparrow$) and AUC-Recall@k~($\uparrow$) of \method{} and SARD-c for different synonym rates for the \textbf{office scene}.
   The synonym rate represents the probability of substituting words using synonym mappings.
   For \method{} the dots represent the average results after running with ten different seeds, and the shaded area visualizes the corresponding standard deviation.
   SARD-c was run only once for each rate as the calculation is deterministic.
   }
   \label{fig:eval-mut-office}
\end{figure}

\subsection{Gaussian Noise Experiments} \label{sec:gauss-noise-abl}
We test robustness to error-prone SGGs by adding Gaussian noise directly to the concatenated word embedding features. 
Specifically, we perturb the clean embedding $\mathbf{t}$ as
\begin{equation}
\mathbf{t}_\text{noisy} = \mathbf{t} + \boldsymbol{\epsilon}, \quad \boldsymbol{\epsilon} \sim \mathcal{N}(0, \sigma^2 I)
\end{equation}
and use $\mathbf{t}_\text{noisy}$ as input to the autoencoder.
The intuition is that a small deviation of the embedding vectors corresponds to a shift towards semantically similar words, thereby simulating variability in object and relationship descriptions.
\cref{tab:sgg-ablation-results}  shows the results for different values of $\sigma$.
As the noise does not yield a consistent improvement, we exclude it from the final model.

\begin{table}[t]
\centering
\caption{Results for experiments for the scene graph generator robustness, with a different scene graph generator and gaussian noise with different $\sigma$. AUROC and AUC-Recall@k on the SARD dataset.}
\resizebox{0.9\linewidth}{!}{
\begin{tabular}{l|l||l|l}
    \toprule
    Scene & Experiment & AUROC ($\uparrow$) & AUC-Recall@k ($\uparrow$)\\
    \midrule
    \multirow{4}{*}{\begin{tabular}[c]{@{}c@{}}Dining\\Room\end{tabular}} & RelTR & 95.57 \scriptsize{$\pm$ 0.88} & 87.29 \scriptsize{$\pm$ 3.2} \\
     & $\sigma = 0.01$ & 97.43 \scriptsize{$\pm$ 0.57} & 93.08 \scriptsize{$\pm$ 1.41}\\
     & $\sigma = 0.05$ & 97.46 \scriptsize{$\pm$ 0.47} & 93.44 \scriptsize{$\pm$ 1.18}\\
     & $\sigma = 0.10$ & 96.11 \scriptsize{$\pm$ 1.22} & 90.37 \scriptsize{$\pm$ 2.54}\\
    \midrule
    \multirow{4}{*}{Office} & RelTR & 96.09 \scriptsize{$\pm$ 1.16} & 89.47 \scriptsize{$\pm$ 3.11}\\
    & $\sigma = 0.01$ & 95.39 \scriptsize{$\pm$ 1.54} & 89.77 \scriptsize{$\pm$ 3.22}\\
    & $\sigma = 0.05$ & 95.60 \scriptsize{$\pm$ 1.46} & 90.24 \scriptsize{$\pm$ 3.05}\\
     & $\sigma = 0.10$ & 92.91 \scriptsize{$\pm$ 1.69} & 84.97 \scriptsize{$\pm$ 3.45}\\
    \bottomrule
\end{tabular}
}
\label{tab:sgg-ablation-results}
\end{table}

\begin{table}[t]
\centering
\caption{Results for experiments with different word embedding models.
All variations were executed with ten different seeds and the average and standard deviation is provided. The best performing model is \textbf{highlighted}.}
\resizebox{\linewidth}{!}{
\begin{tabular}{l|l||l|l}
    \toprule
    Scene & Embedding Model & AUROC ($\uparrow$) & AUC-Recall@k ($\uparrow$)\\
    \midrule
    \multirow{4}{*}{\begin{tabular}[c]{@{}c@{}}Dining\\Room\end{tabular}} & all-MiniLM-L6-v2~\cite{word-embed-miniLM} & 96.34 \scriptsize{$\pm$ 0.69} & 90.98 \scriptsize{$\pm$ 1.35} \\
     & all-mpnet-base-v2~\cite{word-embed-mpnet} & 97.25 \scriptsize{$\pm$ 0.55} & 93.16 \scriptsize{$\pm$ 1.22}\\
     & Qwen3-Embedding-8B~\cite{word-embed-yang2025qwen3} & \textbf{97.99} \scriptsize{$\pm$ 0.58} & \textbf{94.77} \scriptsize{$\pm$ 1.48}\\
     & GloVe~\cite{glove_v1} & 97.85 \scriptsize{$\pm$ 0.6} & 92.85 \scriptsize{$\pm$ 1.52}\\
    \midrule
    \multirow{4}{*}{Office} & all-MiniLM-L6-v2 & 95.43 \scriptsize{$\pm$ 2.30} & 89.68 \scriptsize{$\pm$ 4.81}\\
    & all-mpnet-base-v2 & 95.40 \scriptsize{$\pm$ 2.74} & 89.69 \scriptsize{$\pm$ 5.79}\\
    & Qwen3-Embedding-8B & \textbf{96.09} \scriptsize{$\pm$ 1.71} & \textbf{91.30} \scriptsize{$\pm$ 3.52}\\
    & GloVe & 95.29 \scriptsize{$\pm$ 1.65} & 89.57 \scriptsize{$\pm$ 3.46}\\
    \bottomrule
\end{tabular}
}
\label{tab:sentence-embed-results}
\end{table}

\subsection{Encoding Independence} \label{sec:encoding-independence}
\textbf{Scene Graph Generator Independence.}
To show the independence of \method{} from the chosen scene graph generator, we replaced the modular SGG component with RelTR~\cite{reltr} and provide the results in \cref{tab:sgg-ablation-results}. The similar performance underlines the stability of \method{} regarding different scene graph generators.

\textbf{Word Embedding Independence.}
To demonstrate independence from the chosen word embedding model, we conducted experiments substituting GloVe~\cite{glove_v1} with stronger sentence embedding models, which encode complete sentences rather than individual words.
For these models, each triplet is encoded as a full sentence of the form: \texttt{`In a \{scene\_type\}, \{obj1\} \{relation\} \{obj2\}'}, incorporating both the scene context and the triplet components.
The results are provided in \cref{tab:sentence-embed-results}.
Despite being a simpler word-level model, GloVe achieves competitive performance compared to the more computationally expensive sentence embedding models, justifying its use in \method{}.

\section{Additional Design Study Results}
\Cref{tab:ablation-office} and \cref{fig:eval-latent-auc} show the office scene and AUC-Recall@k counterparts to the design study results presented in \cref{sec:design-study}.

\begin{figure}[t]
  \centering
  \includegraphics[width=0.8\linewidth]{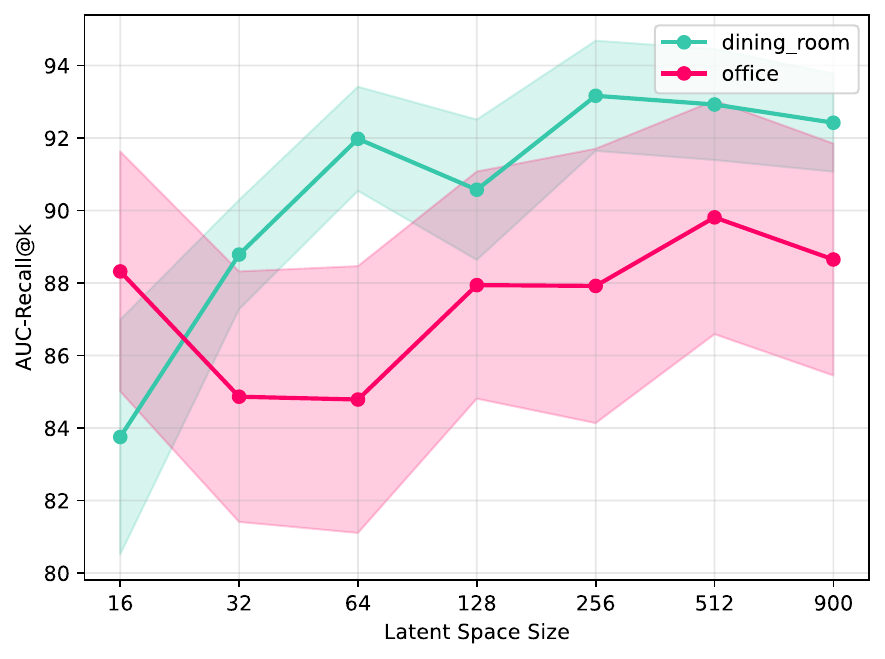}
   \caption{Ablation results with the AUC-Recall@k ($\uparrow$) for different latent space dimensions of the autoencoder.}
   \label{fig:eval-latent-auc}
\end{figure}

\begin{table}[t]
\centering
\caption{Ablation study results for \textbf{office} scene. All experiments were executed with ten different seeds and the average results are listed with their standard deviation. $d_z$ is the input dimension to the normalizing flow.}
\label{tab:ablation-office}
\resizebox{\linewidth}{!}{
\begin{tabular}{l|l||c|c}
    \toprule
    Experiment & $d_z$ & AUROC ($\uparrow$) & AUC-Recall@k ($\uparrow$)\\
    \midrule
    Feature Sum & 128 & 94.18 \scriptsize{$\pm$ 2.45} & 87.35 \scriptsize{$\pm$ 5.23}\\
    Feature Mult & 128 & 94.65 \scriptsize{$\pm$ 3.18} & 88.24 \scriptsize{$\pm$ 6.74}\\
    Node Only & 512 & 90.89 \scriptsize{$\pm$ 4.33} & 80.16 \scriptsize{$\pm$ 9.26}\\
    No AE & 900 & 85.74 \scriptsize{$\pm$ 2.98} & 71.82 \scriptsize{$\pm$ 5.18} \\
    \method{} (ours) & 512 & \textbf{95.29} \scriptsize{$\pm$ 1.65} & \textbf{89.57} \scriptsize{$\pm$ 3.46}\\
    \bottomrule
\end{tabular}
}
\end{table}

\section{Qualitative Analysis} \label{sec:qualitative-analysis}
To better understand the strengths and weaknesses of \method{}, we conduct a qualitative analysis across both scenes.
True positives include unusual object placements such as `banana-on-chair' or `shoe-on-desk'.
The false negative `pillow-on-table' scored near the decision threshold, indicating a genuinely ambiguous relationship, likely due to rare training occurrences.
A notable false positive is the triplet `flower-under-clock' which is detected by the scene graph generator despite not being in the image.
However the detection of this triplet as anomalous is correct.

\Cref{fig:sgs-w-score} shows subsets of the generated scene graphs with normalized triplet anomaly scores, with the two highest-scoring triplets highlighted.
In \cref{fig:sg_dining_room_score}, `flower-on-table' correctly receives the highest anomaly score.
In \cref{fig:sg_office_score}, however, `bag-on-chair' scores highest while the true anomaly `shoe-on-chair' ranks second, suggesting that bags on chairs were rarely seen during training, causing an elevated anomaly score.

\section{Example Scene Graphs of Images} \label{sec:img-sg-example}
We present example images from the dataset with their corresponding top 30 triplets.
Three examples each from the dining room and office scenes can be found in \cref{fig:images}.
The corresponding scene graphs for the dining room (see \cref{fig:sg_dining_room}) and office (see \cref{fig:sg_office}) examples are also presented.

%%%%%%%% Example Images %%%%%%%%%%%%%%

\begin{figure*}[ht]
  \centering
  \begin{subfigure}[t]{0.48\linewidth}
    \centering
    \includegraphics[width=0.9\linewidth]{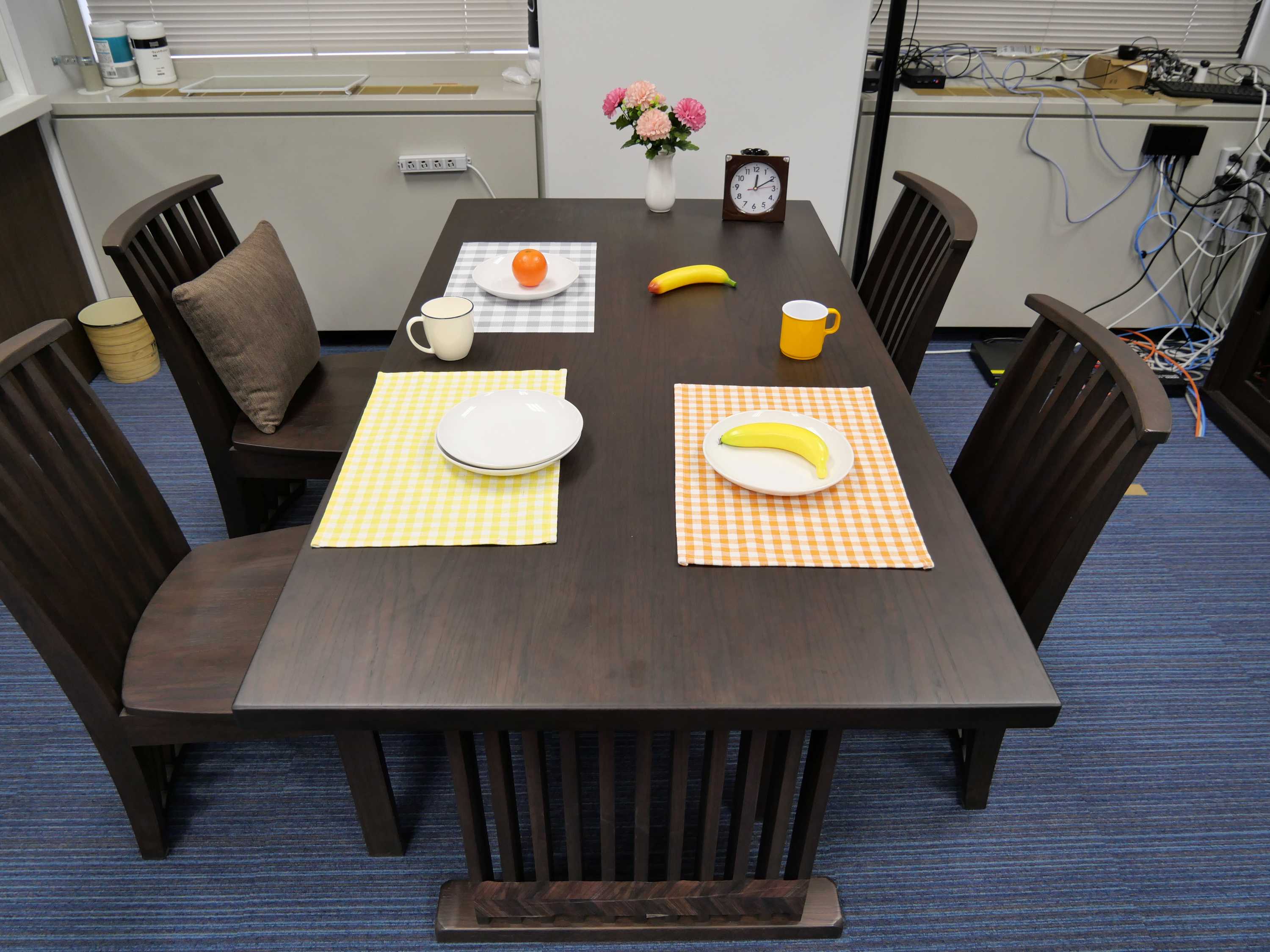}
    \caption{Dining room example 1}
    \label{fig:dining_a}
    
    \vspace{1em}
    \includegraphics[width=0.9\linewidth]{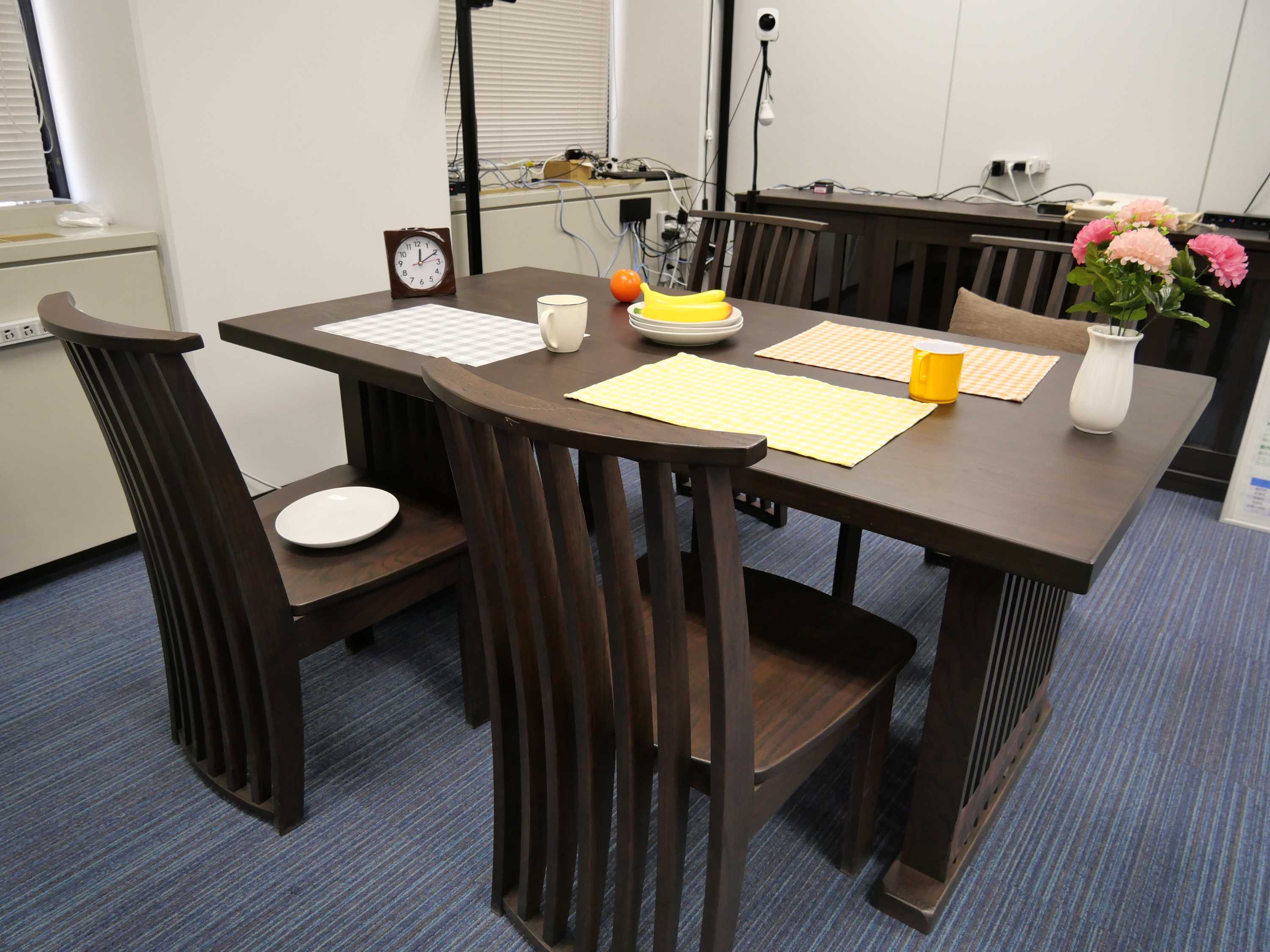}
    \caption{Dining room example 2}
    \label{fig:dining_b}
    
    \vspace{1em}
    \includegraphics[width=0.9\linewidth]{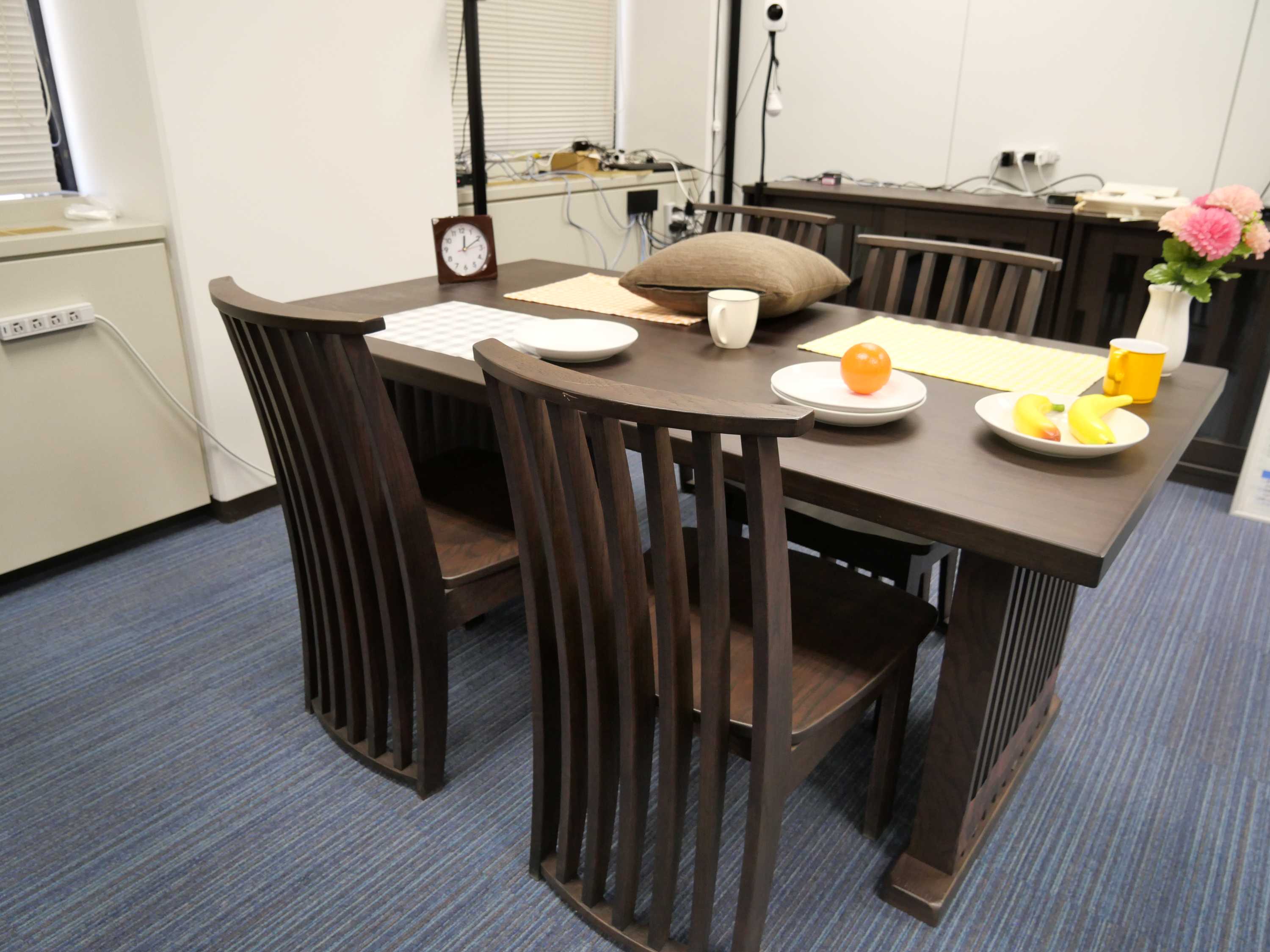}
    \caption{Dining room example 3}
    \label{fig:dining_c}
  \end{subfigure}
  \hfill
  \begin{subfigure}[t]{0.48\linewidth}
    \centering
    \includegraphics[width=0.9\linewidth]{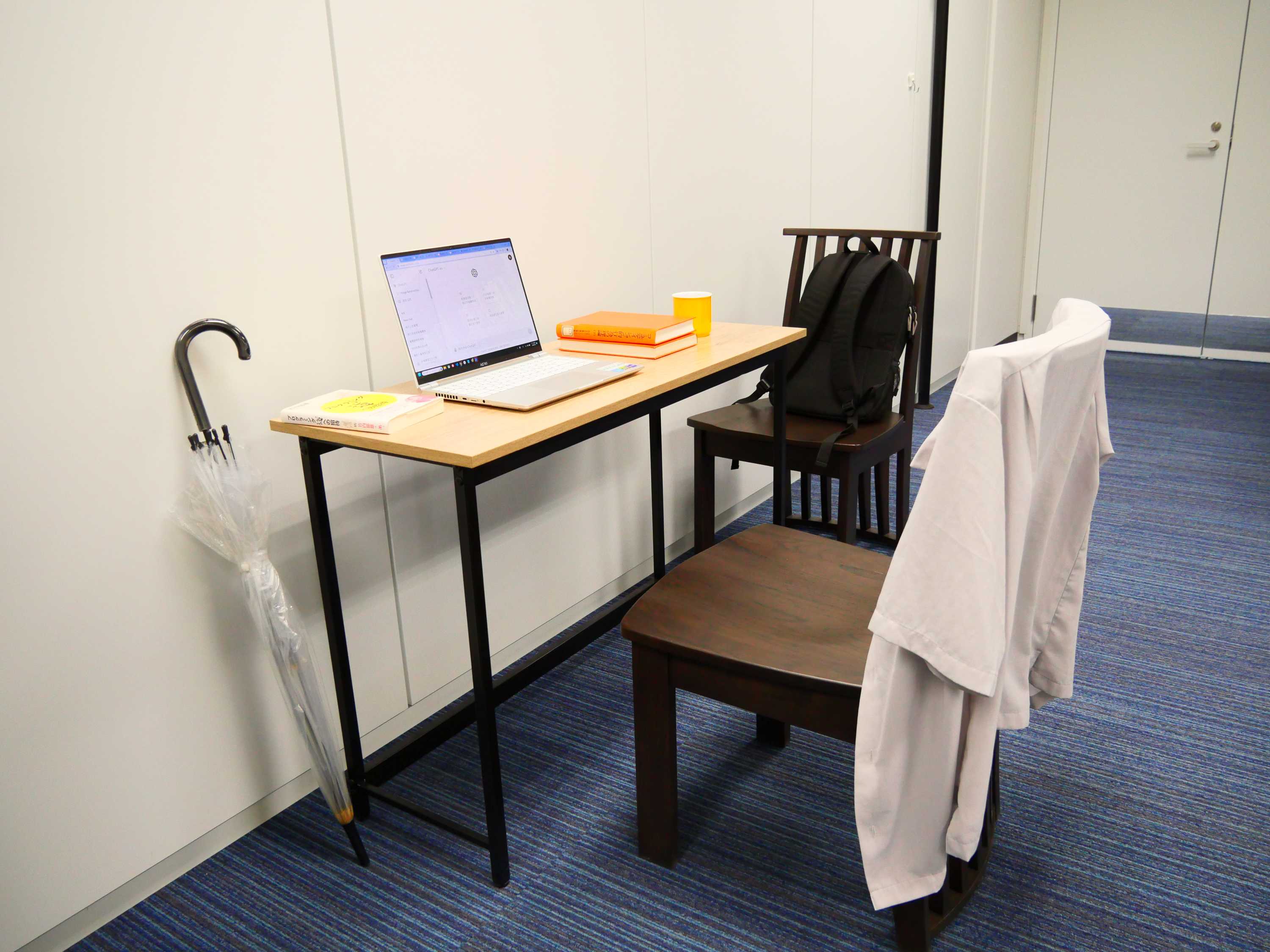}
    \caption{Office example 1}
    \label{fig:office_a}
    
    \vspace{1em}
    \includegraphics[width=0.9\linewidth]{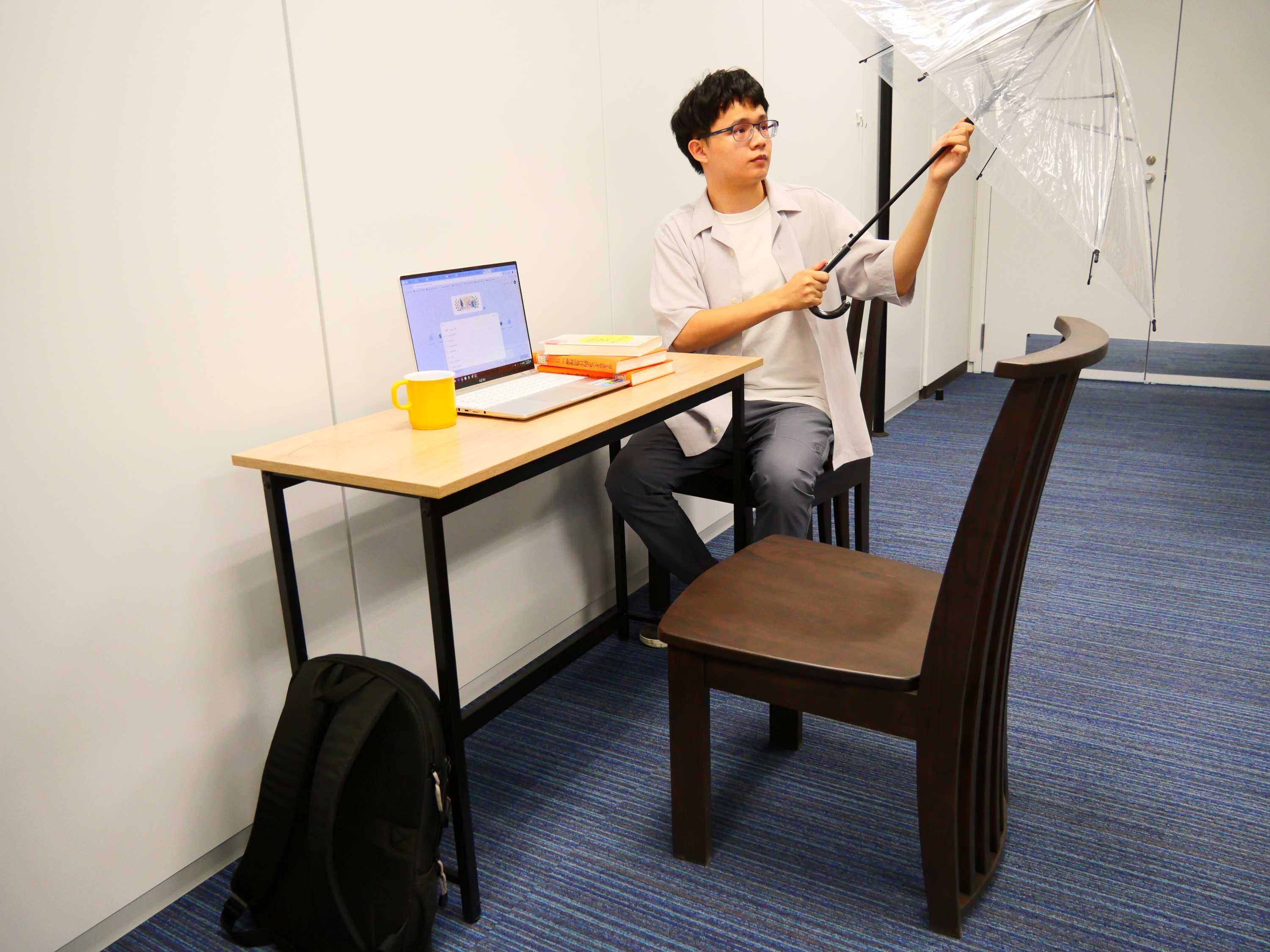}
    \caption{Office example 2}
    \label{fig:office_b}
    
    \vspace{1em}
    \includegraphics[width=0.9\linewidth]{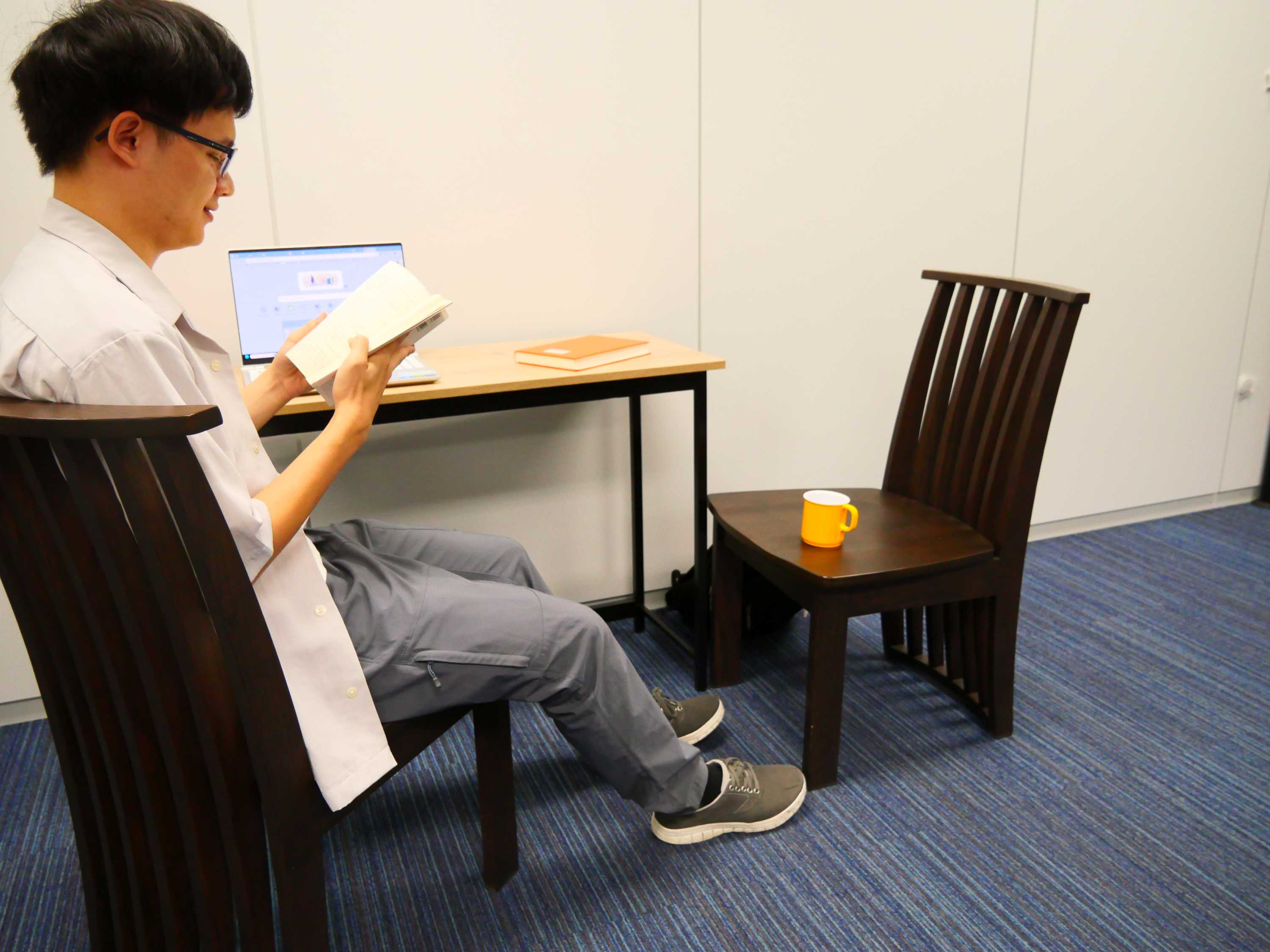}
    \caption{Office example 3}
    \label{fig:office_c}
  \end{subfigure}
  \caption{Example images from \textbf{dining room} (left) and \textbf{office} (right). The top images (a) and (d) are normal, while the others contain anomalies.}
  \label{fig:images}
\end{figure*}

%%%%%%%%%%%%%% Scene Graphs %%%%%%%%%%%%%%%%%%%%

\begin{figure*}[ht]
  \centering
  \begin{subfigure}[b]{0.7\linewidth}
    \includegraphics[width=\linewidth]{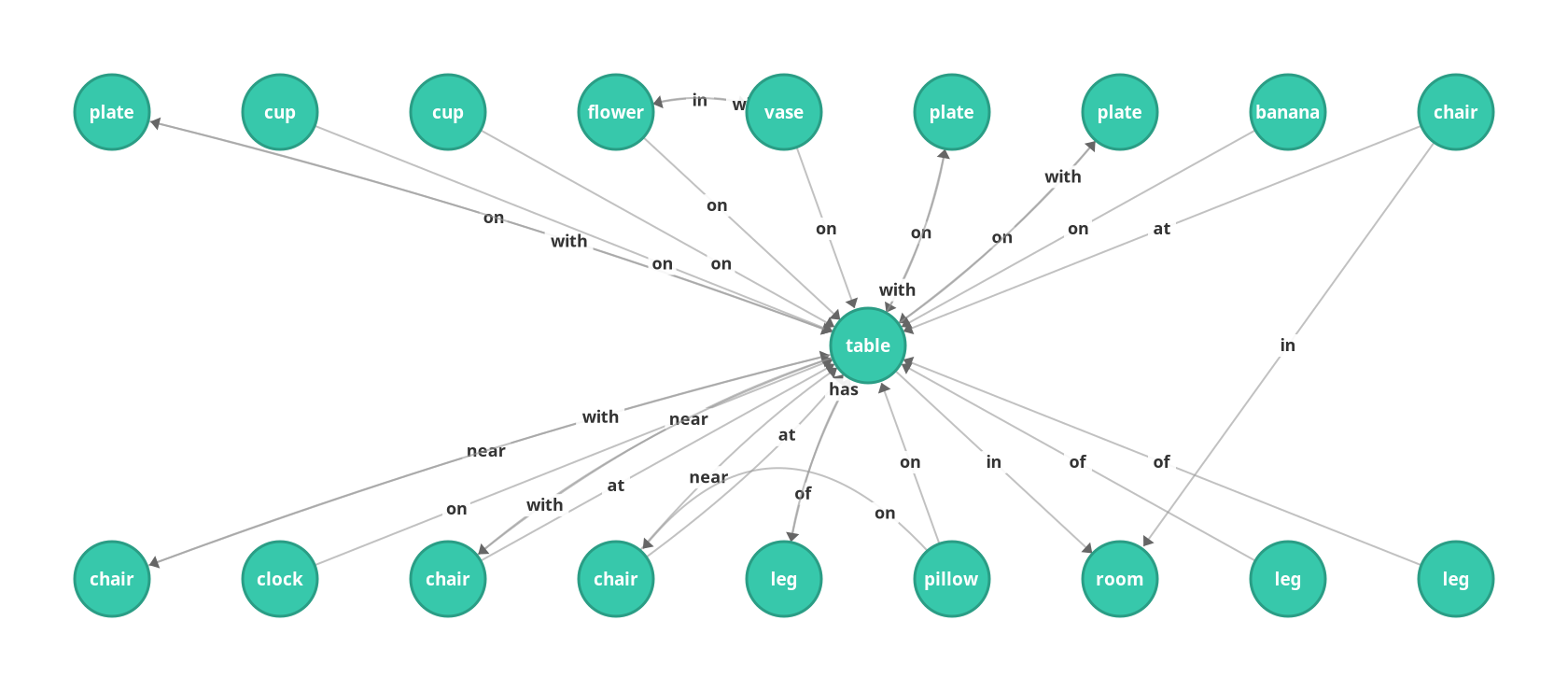}
    \caption{Scene graph dining room example 1}
    \label{fig:sg_dining_a}
  \end{subfigure}
  
  \begin{subfigure}[b]{0.6\linewidth}
    \includegraphics[width=\linewidth]{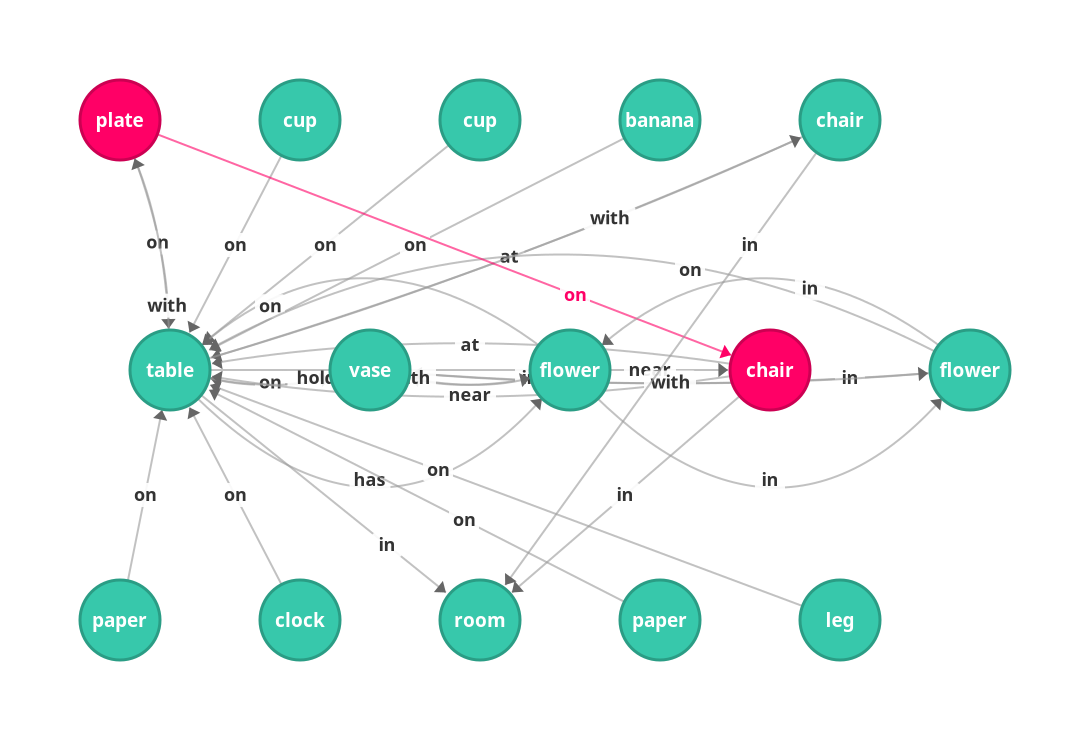}
    \caption{Scene graph dining room example 2}
    \label{fig:sg_dining_b}
  \end{subfigure}
  
  \begin{subfigure}[b]{0.7\linewidth}
    \includegraphics[width=\linewidth]{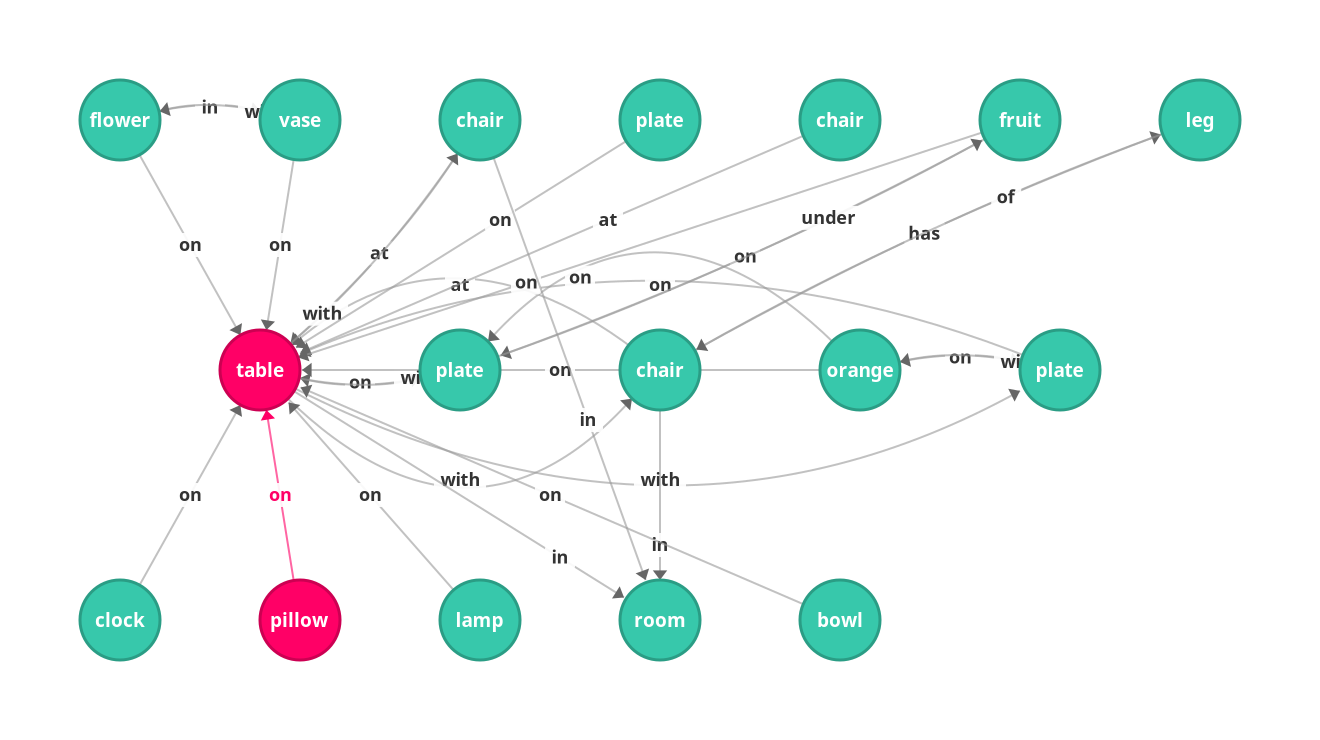}
    \caption{Scene graph dining room example 3}
    \label{fig:sg_dining_c}
  \end{subfigure}
  
  \caption{Example scene graphs from \textbf{dining room} images, constructed from the top 30 triplets. The last two scene graphs contain anomalies, which are highlighted red.}
  \label{fig:sg_dining_room}
\end{figure*}

\begin{figure*}[ht]
  \centering
  \begin{subfigure}[b]{0.7\linewidth}
  \addtocounter{subfigure}{3}
    \includegraphics[width=\linewidth]{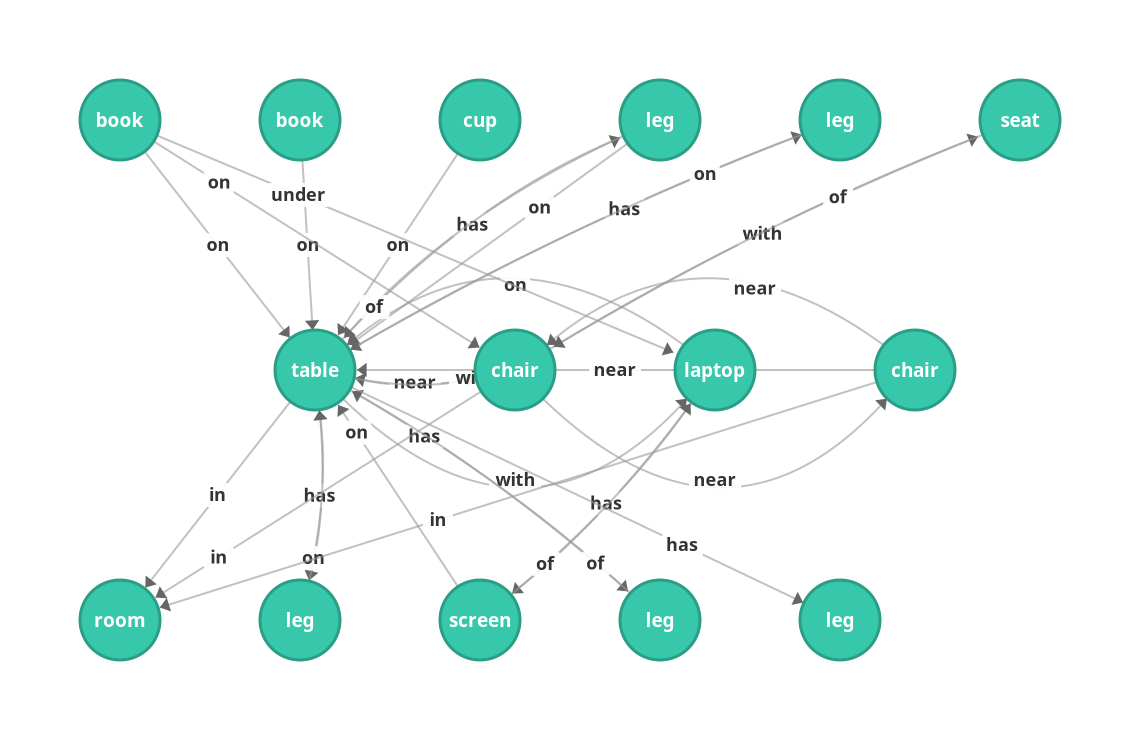}
    \caption{Scene graph office example 1}
    \label{fig:sg_office_a}
  \end{subfigure}
  
  \begin{subfigure}[b]{0.7\linewidth}
    \includegraphics[width=\linewidth]{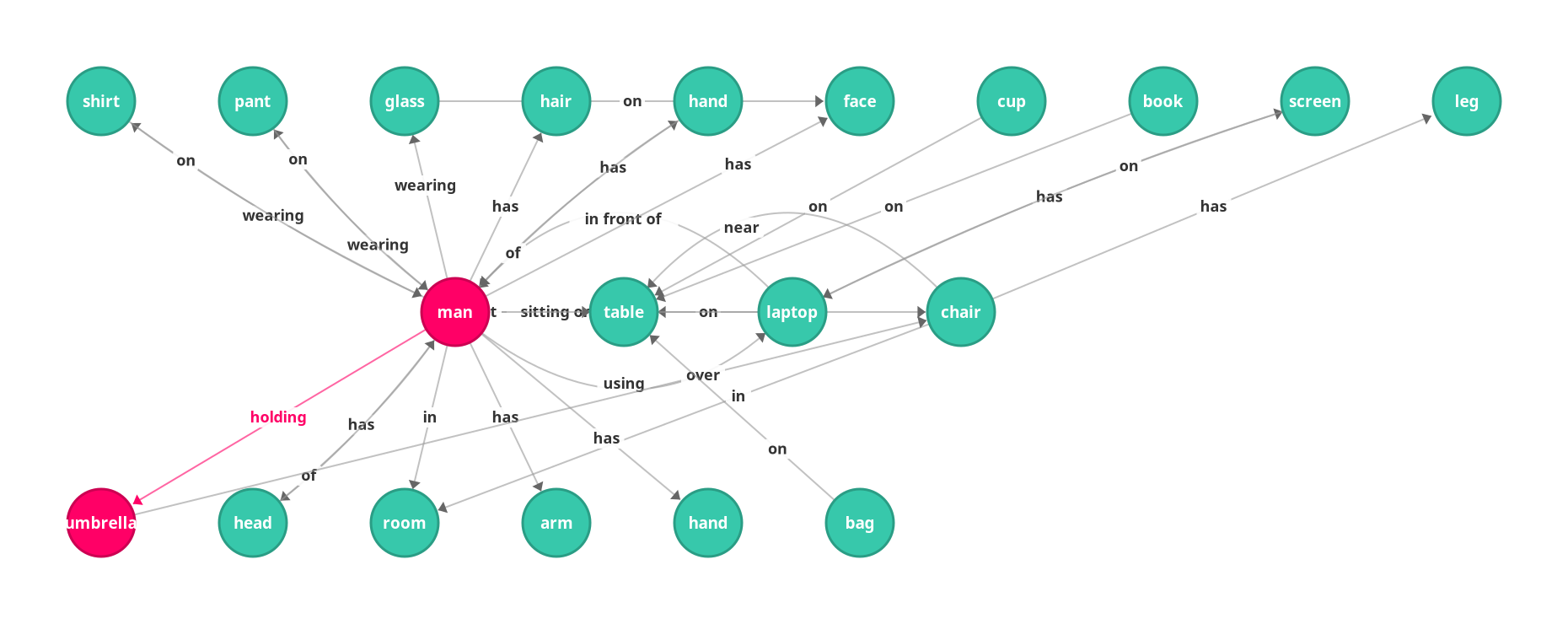}
    \caption{Scene graph office example 2}
    \label{fig:sg_office_b}
  \end{subfigure}
  
  \begin{subfigure}[b]{0.7\linewidth}
    \includegraphics[width=\linewidth]{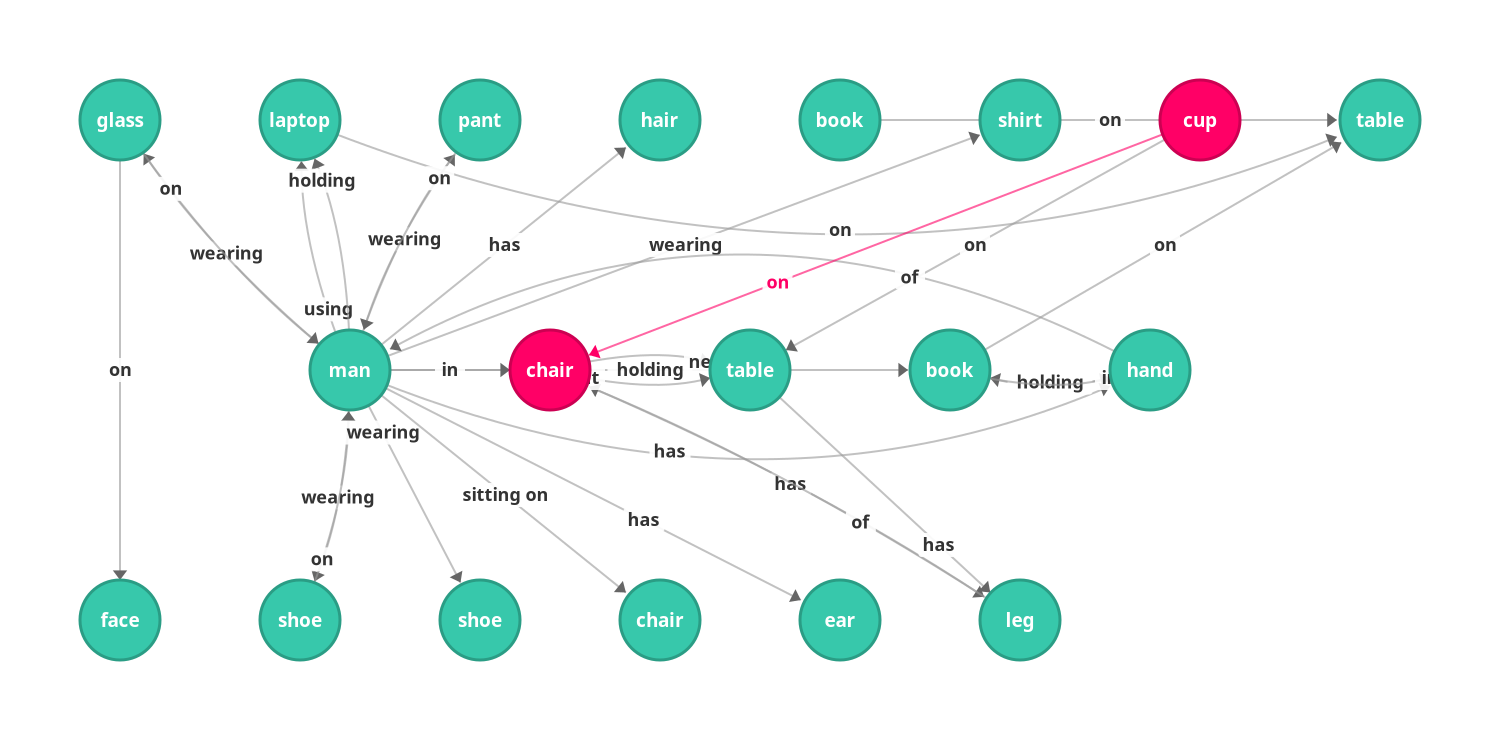}
    \caption{Scene graph office example 3}
    \label{fig:sg_office_c}
  \end{subfigure}
  
  \caption{Example scene graphs from \textbf{office} images, constructed from the top 30 triplets. The last two scene graphs contain anomalies, which are highlighted red.}
  \label{fig:sg_office}
\end{figure*}

\begin{figure*}[b]
  \centering
  \begin{subfigure}[b]{0.65\textwidth}
    \includegraphics[width=\linewidth]{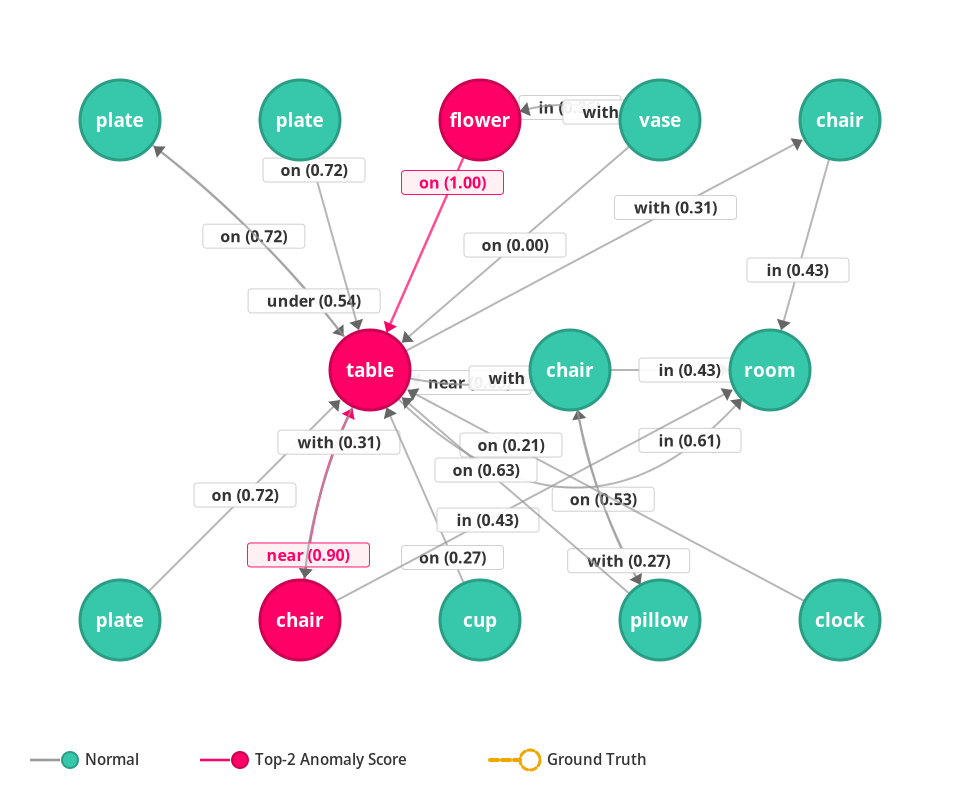}
    \caption{Scene graph dining room example with normalized anomaly scores}
    \label{fig:sg_dining_room_score}
  \end{subfigure}
  \begin{subfigure}[b]{0.6\textwidth}
    \includegraphics[width=\linewidth]{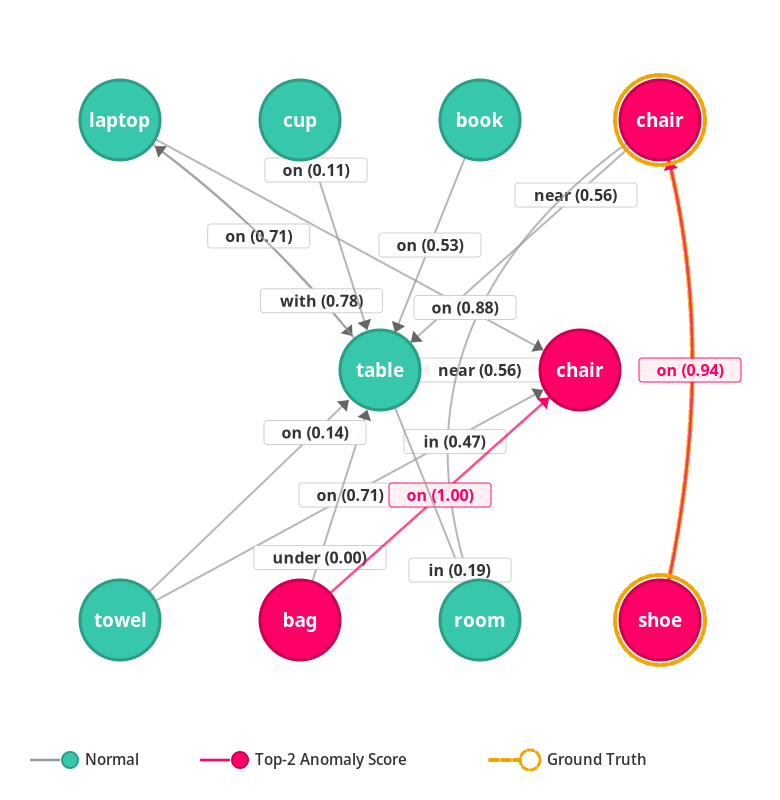}
    \caption{Scene graph office example with normalized anomaly scores}
    \label{fig:sg_office_score}
  \end{subfigure}

   \caption{Example subsets of scene graphs with normalized anomaly scores. The red highlighted elements belong to the two triplets with the highest anomaly score and the ground truth is marked with an additional, yellow circle.}
   \label{fig:sgs-w-score}
\end{figure*}

\end{document}